\documentclass[acmsmall]{acmart}
\usepackage{CJKutf8}
\pdfoutput=1
\AtBeginDocument{%
  \providecommand\BibTeX{{%
    \normalfont B\kern-0.5em{\scshape i\kern-0.25em b}\kern-0.8em\TeX}}}

\setcopyright{acmlicensed}
\acmJournal{CSUR}
\acmYear{2024} \acmVolume{1} \acmNumber{1} \acmArticle{1} \acmMonth{1}\acmDOI{10.1145/3674501}

\usepackage{multirow}
\usepackage{lscape}%isn't allowed
\usepackage[figuresright]{rotating}
\usepackage{float}
\usepackage[T1]{fontenc} % 参考文献的作者姓名字母上带有变音符号
\usepackage{tabularx} % automatic line-break
\usepackage{blindtext}
\usepackage{filecontents}
\usepackage{pgfplotstable}
\pgfplotsset{compat=1.12}
\usetikzlibrary{patterns}
\usepgfplotslibrary{groupplots}
\usepackage{makecell} % 表格内换行
\usepackage{graphicx} %表格缩小功能
\usepackage{booktabs} %给表格画横线
\usepackage{bm} % 数学公式中加粗字体
\usepackage{bbding} %特殊符号 对错
\usepackage{hyperref} % url超链接
\usepackage{lineno} % 加行号

\usepackage{tikz}
\usepackage[edges]{forest}
\definecolor{hiddendraw}{RGB}{205, 44, 36}
\definecolor{hidden-blue}{RGB}{194,232,247}
\definecolor{hidden-orange}{RGB}{243,202,120}
\definecolor{hidden-yellow}{RGB}{242,244,193}

\usepackage{soul}

\usepackage[normalem]{ulem}
\useunder{\uline}{\ul}{}

\acmJournal{CSUR}
\begin{document}
% \setcopyright{acmlicensed}
% \acmJournal{CSUR}
% \acmYear{2024} \acmVolume{1} \acmNumber{1} \acmArticle{1} \acmMonth{1}\acmDOI{10.1145/3674501}

% \linenumbers % 编号开始
% https://tex.stackexchange.com/questions/365752/how-to-remove-acm-reference-format-box-in-sig-conf-template
\settopmatter{printacmref=false}

\title{A Comprehensive Survey on Relation Extraction:  Recent Advances and New Frontiers}


\author{Xiaoyan Zhao}
\email{xzhao@se.cuhk.edu.hk}
\affiliation{%
  \institution{The Chinese University of Hong Kong}
  \country{China}
}

\author{Yang Deng}
\email{ydeng@smu.edu.sg}
\affiliation{%
  \institution{Singapore Management University}
  \country{Singapore}
}

\author{Min Yang}
\authornote{Min Yang and Ying Shen are corresponding authors.}
\email{min.yang@siat.ac.cn}
\affiliation{%
  \institution{Shenzhen Institutes of Advanced Technology, Chinese Academy of Sciences}
  \country{China}
}

\author{Lingzhi Wang}
\email{lzwang1120@gmail.com}
\affiliation{%
  \institution{The Chinese University of Hong Kong}
  \country{China}
}

\author{Rui Zhang}
\email{rayteam@yeah.net}
\affiliation{%
  \institution{Huazhong University of Science and Technology (www.ruizhang.info)}
  \country{China}
}

\author{Hong~Cheng}
\email{hcheng@se.cuhk.edu.hk}
\affiliation{%
  \institution{The Chinese University of Hong Kong}
  \country{China}
}

\author{Wai~Lam}
\email{wlam@se.cuhk.edu.hk}
\affiliation{%
  \institution{The Chinese University of Hong Kong}
  \country{China}
}

\author{Ying Shen}
\authornotemark[1]
\email{sheny76@mail.sysu.edu.cn}
\affiliation{%
  \institution{Sun Yat-Sen University}
  \country{China}
}

\author{Ruifeng~Xu}
\email{xuruifeng@hit.edu.cn}
\affiliation{%
  \institution{Harbin Institute of Technology (Shenzhen)}
  \country{China}
}

\iffalse
\author{Xiaoyan~Zhao$^1$, Yang~Deng$^1$, Min~Yang$^2$, Rui~Zhang$^3$, Hong~Cheng$^1$, Wai~Lam$^1$, Ying Shen$^4$, Ruifeng~Xu$^5$}
\affiliation{%
  \institution{$^1$The Chinese University of Hong Kong, 
    $^2$SIAT, Chinese Academy of Sciences, ~ $^3$Tsinghua University, ~ $^4$Sun Yat-Sen University, China, ~$^5$Harbin Institute of Technology (Shenzhen)}
    \country{China}}
\email{{xzhao, ydeng, hcheng, wlam}@se.cuhk.edu.hk, min.yang@siat.ac.cn, rayteam@yeah.net, xuruifeng@hit.edu.cn}
\fi

\renewcommand{\shortauthors}{Xiaoyan Zhao, et al.}

\begin{abstract}
Relation extraction (RE) involves identifying the relations between entities from underlying content. RE serves as the foundation for many natural language processing (NLP) and information retrieval applications, such as knowledge graph completion and question answering. In recent years, deep neural networks have dominated the field of RE and made noticeable progress. Subsequently, the large pre-trained language models have taken the state-of-the-art RE to a new level. This survey provides a comprehensive review of existing deep learning techniques for RE. First, we introduce RE resources, including datasets and evaluation metrics. Second, we propose a new taxonomy to categorize existing works from three perspectives, i.e., text representation, context encoding, and triplet prediction. Third, we discuss several important challenges faced by RE and summarize potential techniques to tackle these challenges. Finally, we outline some promising future directions and prospects in this field. This survey is expected to facilitate researchers' collaborative efforts to address the challenges of real-world RE systems.
\end{abstract}

%%
%% The code below is generated by the tool at .
% https://dl.acm.org/ccs

% \ccsdesc[500]{General and reference~Surveys and overviews}


\maketitle

%\tableofcontents
%\newpage

\section{Introduction} 
	\label{survey-intro}
Relation extraction (RE) is an essential task in natural language processing (NLP), which involves extracting entities and relations between them from underlying content. In this paper, we primarily focus on binary relations as the main unit of analysis for RE tasks. Each relation is represented as a triplet  \textit{$\langle$head$\_$entity, relationship, tail$\_$entity$\rangle$}, consisting of two entities and the relation between them. 
RE facilitates the extraction of structured information from vast troves of unstructured texts, thereby unlocking the value hidden within such data. 
It can be used for many downstream applications \cite{Nayak2021DeepNA,xu2022conreader}, such as knowledge graph completion \cite{chen2022hybrid} and alignment \cite{zhang2022benchmark}, question answering \cite{luoKnowledgeBaseQuestion2018}, and information retrieval \cite{yang2020biomedical}. 
In the era of Large Language Models (LLMs), RE methods continue to demonstrate significant advantages. LLMs struggle to accurately retain all the knowledge implied within the text, especially in handling long-tail texts where errors in judging relationships between entities are prone to occur. Thus, RE techniques serve as a potent technical complement in enhancing the accuracy of LLMs. 
Furthermore, in rapidly evolving domains where new entities, relationships, and concepts frequently emerge, RE methods offer the flexibility to effectively adapt to and incorporate new information, offering scalable solutions to the daunting task of mining structured insights from the vast expanse of unstructured data.
Therefore, designing automatic approaches to extract the relations between entities contained in unstructured texts becomes increasingly necessary, leading to the booming development of RE.

\begin{figure}[t] 
	\centering 
	\includegraphics[width=0.45 \textwidth]{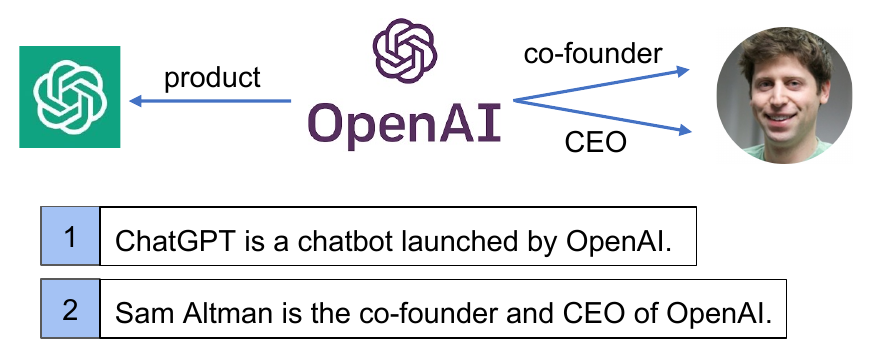} 
	\caption{Examples of relation extraction.}
	\label{fig:intro_example} 
\vspace{-0.4cm}
\end{figure}

In recent years, advances in deep neural networks (DNNs) and pre-trained language models (PLMs) have significantly improved the performance of RE. These approaches can be categorized into two main types: \textbf{the pipeline-based RE approaches} \cite{miwa2016end} and \textbf{joint RE approaches} \cite{yuan2021relation,zhao2021asking,nayak2020effective}.
Pipeline-based approaches extract entities and relations from unstructured text through two separate stages, which first identify entities from the text and then detect the relation between any pairs of entities.  For example, as illustrated in Figure \ref{fig:intro_example}, given the sentence ``\textit{ChatGPT is a chatbot launched by OpenAI}'', pipeline-based approaches first identify the entities ``\textit{ChatGPT}'' and ``\textit{OpenAI}'', and then predict the relation ``\textit{product}'' between the two entities.
In the early stage, pipeline-based RE approaches \cite{huang2015bidirectional,miwa2016end,Pawar2017RelationE} primarily use Named Entity Recognition (NER) tools to extract entities, and then classify the relations of entity pairs using supervised learning algorithms with feature engineering.
Pipeline-based RE methods \cite{wu2019improving,Bhartiya2021DiSReXAM,chen2022knowprompt} often assume that the target entities are already identified, and the RE models merely need to predict the relations between any pair of entities.
However, because the entity and relation extraction processes are separated, pipeline-based approaches tend to suffer from error propagation, where relation classification can be affected by errors introduced during entity recognition. 

Joint (non-pipeline) approaches, on the other hand, aim to address this challenge by jointly modeling entity recognition and relation classification tasks within a unified framework. Taking the second example in Figure \ref{fig:intro_example} as an example, the sentence ``\textit{Sam Altman is the co-founder and CEO of OpenAI}''  contains two relationships (i.e., ``\textit{co-founder}'' and ``\textit{CEO}'') with overlapping entities. RE systems must be able to accurately identify and distinguish between overlapping entities and relationships. Joint RE approaches tend to be less susceptible to error propagation due to several key reasons. First, this holistic training approach allows the model to learn optimal representations for both tasks concurrently, minimizing the impact of errors in one aspect on the other. Second, joint models can optimize objectives directly related to the overall task, such as maximizing the likelihood of correct entity pairs and relations. Third, by jointly learning entity recognition and relation extraction tasks, joint models can adapt to errors in one task by leveraging information from the other, thus compensating for potential mistakes made at earlier stages. 
So far, many joint RE approaches have been proposed to extract entity and relation simultaneously. We generally divide them into four categories: span-based approaches \cite{zhao2022exploring,yan2022empirical}, sequence-to-sequence (Seq2Seq) approaches \cite{zeng2020copymtl}, MRC-based approaches \cite{li2019entity,zhao2021asking}, and sequence labeling approaches \cite{yuan2021relation,hillebrand2022kpi}. 

Despite the advances of deep learning for RE, several challenging problems still need to be solved in real-world scenarios. For example, many relations are ``long-tail'', where only a few frequent relations receive sufficient training examples. In contrast, the remaining infrequent relations usually suffer from a lack of labeled training data. However, deep learning requires massive amounts of training corpus, which is difficult to obtain in many real-life applications, especially in \textit{low-resource settings}. \textbf{Distant supervision relation extraction (DSRE)} \cite{yuan2019cross} is particularly appealing as it leverages existing structured information, such as knowledge graphs (KGs) and databases, to generate labeled training data automatically. Nevertheless, distant supervision techniques may suffer from the wrong labeling problem and fail to handle long-tail relations with limited labeled instances. 
Therefore, \textbf{few-shot relation extraction (FSRE)} with limited labeled training samples has become a hot research topic \cite{Qu2020FewshotRE,chen2022knowprompt}. 

In addition, most existing studies focus on extracting relational facts from individual sentences. However, many real-life applications require the RE systems to identify entities and relations from a long document with multiple sentences. Following this direction, some recent studies~\cite{christopoulou2019connecting,Lee2021GraphBN} have been proposed to solve \textit{cross-sentence RE}, which attempt to identify relations that are mentioned across multiple sentences. Generally, there are two main research lines on cross-sentence RE. 
The first line of research is \textbf{document-level RE} \cite{jia2019document}, which has the potential to overcome the inherent limitations of sentence-level approaches and better capture the full range of relational information present in a document. The second line of research is \textbf{dialogue RE} \cite{chen2022knowprompt,Son2022GRASPGM}, which aims to discover relation triplets appearing in multi-turn dialogues. 
Furthermore, the prosperity of RE in the general field motivates some works to focus on domain-specific relation extraction from specialized articles. To the best of our knowledge, domain-specific RE approaches are still under-explored in previous surveys. In this paper, we summarize the advanced RE approaches in specific domains (e.g., scientific, finance, medical, and biochemical). 

Recently, some studies have also focused on other promising yet challenging RE problems, including
\textbf{multi-modal relation extraction} \cite{zheng2021multimodal},
\textbf{cross-lingual relation extraction} \cite{seganti2021multilingual}, \textbf{temporal relation extraction} \cite{tan2021extracting}, and \textbf{evolutionary relation extraction} \cite{zhao2022consistent}.
To facilitate building a comprehensive understanding of RE, we also review recent advances that address these challenging RE problems.

Overall, RE studies
\cite{detroja2023survey,Xu2022TowardsRL}
have been thriving in recent years. Although  there have been several surveys on RE, they do not provide sufficient reviews of the above recent deep learning (DL)-based advances, current challenges, and future directions. In particular, the early surveys \cite{Zelenko2002KernelMF, Bach2007ARO, Pawar2017RelationE} emphasized traditional RE approaches (i.e., rule-based and machine learning-based approaches) in sentence-level settings.  
\citeauthor{detroja2023survey} \shortcite{detroja2023survey} focused on both traditional RE approaches and DL-based approaches, but did not fully explore and omit the recent DL approaches. 
Moreover, 
\citeauthor{Han2020MoreDM} \shortcite{Han2020MoreDM} 
reviewed existing relation extraction approaches from four specific directions (i.e., utilizing more data, performing more efficient learning, handling more complicated context, and orienting more open domains). 
\citeauthor{Xu2022TowardsRL} \shortcite{Xu2022TowardsRL} 
focused on the low-resource RE problem. 
 \citeauthor{Bassignana2022WhatDY} \shortcite{Bassignana2022WhatDY} 
discussed RE datasets and scientific relation classification approaches. 
The most related survey to ours was proposed by 
\citeauthor{Nayak2021DeepNA} \shortcite{Nayak2021DeepNA}, which collectively introduced the general DL-based RE model architectures. 
However, several challenging issues and new frontiers in recent RE studies have not been discussed. 
Specifically, we argue that existing surveys mainly focused on limited aspects of RE and lack an in-depth sorting of the logical relationships among the classic approaches. Moreover, many emerging developments in this field have not yet been adequately explored. 
For example, PLMs and LLMs (e.g., BERT~\cite{devlin2018bert}, GPT-3~\cite{brown2020language} and ChatGPT \footnote{\url{https://chat.openai.com/}}), which have been widely applied to enhance the outcomes of downstream RE in various scenarios, remain largely unexplored in previous RE surveys.
In this survey, we first organize the general frameworks in the representative RE approaches and fully comb the recent studies into categories, illustrating the differences and connections between RE subtasks. 
In addition, we discuss the performance of relation extraction on current solutions in diverse challenging settings (i.e., low-resource settings and cross-sentence settings) and specific domains (i.e., biomedical, finance, legal, and scientific fields), respectively.
Furthermore, we present in-depth analyses that reveal the primary issues of RE with PLMs and discuss four main challenges (multi-modal RE, cross-lingual RE, temporal RE, and evolutionary RE) that need to be addressed.
The holistic and multi-faceted views of RE methods discussed in our survey would allow readers to obtain a comprehensive landscape of  available RE solutions and a good understanding of potential future directions. 
It is worth noting that to provide a comprehensive overview of relation extraction, we selected bibliographical references based on criteria such as the significance of contributions (novel methods, datasets, metrics) and diversity of approaches. This approach ensures the inclusion of influential and representative works, reflecting the latest advancements and trends in the field across various techniques, datasets, and applications.

\textbf{Contributions of this survey.} 
This survey aims to provide a comprehensive overview of deep learning techniques in RE, which can provide researchers and practitioners with a comprehensive landscape of this area. First, we introduce representative RE datasets for verifying the RE methods. Second, we present a taxonomy  classifying the representative RE approaches into several categories. 
Moreover, we explore and summarize the recent challenges and solutions faced by RE. 
Lastly, we present potential future directions in this field. This survey serves to facilitate
collaborative efforts among researchers in tackling the challenges of relation extraction.

In summary, we offer a comprehensive survey of RE techniques, analyzing the performance of RE models across various task settings and summarizing the limitations of existing models along with future directions for development. We begin with an overview of the research area, emphasizing the existing gaps in the literature concerning RE. Moving forward, Section~\ref{sec:preliminary} delves into datasets and evaluation metrics, establishing a foundational understanding for subsequent discussions. In Section~\ref{sec:dl_techniques_for_RE}, we explore deep learning techniques tailored for RE, with a taxonomy of text representation, context encoding, and triplet prediction. Section~\ref{sec:challenging_RE_problems_and_solutions} discusses current challenging RE problems and solutions, including handling low-resource scenarios, cross-sentence extraction, and adapting to domain-specific RE. Subsequently, in Section~\ref{sec:PLMs}, we critically examine the integration of pre-trained language models and propose future directions, encompassing multi-modal, cross-lingual, temporal, evolutionary, and explainable RE approaches. Finally, in Section~\ref{sec:future_directions}, we conclude the paper by summarizing key findings and outlining promising directions for further research.

\section{Preliminary}
\label{sec:preliminary}
In this section, we first provide a formal problem definition of RE. Then, we introduce the recent benchmark corpora proposed for training deep RE models\footnote{There are several online platforms for RE, including Google Cloud Natural Language~\url{ https://cloud.google.com/natural-language },
IBM Watson Natural Language Understanding~\url{https://www.ibm.com/products/natural-language-understanding} and TextRazor ~\url{https://www.textrazor.com/.}}
. Finally, we present the evaluation metrics for evaluating the RE models.  

\subsection{Problem Definition} 
	\label{survey-problem-definition}
 RE aims to automatically identify the relations between entities in unstructured texts. 
Formally, given a natural language text $x$, the goal of the RE task is to predict a set of triplets, each consisting of a head entity $e_1$, a relation type $r$, and a tail entity $e_2$. 
The entities $e_1$ and $e_2$ can be words, phrases, or other syntactic units in the text, while the relation type $r$ is a predefined type $r \in R$ that describes the relation between $e_1$ and $e_2$.

\begin{table}[]
\caption{Statistics on relation extraction datasets. Document-level datasets are marked with $\star$, while others are sentence-level datasets. The domains of the datasets are divided into general, specific and multi-lingual categories. The $\textbf{\CheckmarkBold}$ mark in the leaderboard column indicates that the dataset has a leaderboard on the Papers with Code website$^2$. }
\label{tab:datasets}
\centering
\small
\scalebox{0.97}{
\resizebox{\textwidth}{!}{%
\begin{tabular}{lllllll}
\toprule
% \rowcolor[HTML]{FFFFFF} 
\textbf{Corpus   Name}         & \textbf{General} & \textbf{Specific} & \textbf{Multi-lingual} & \textbf{Relation} & \textbf{Train/Test} & \textbf{Leaderboard} \\ \midrule
% CoNLL04   \cite{roth2004linear}            & General                & 5     & 1.1k/288            \\
% ACE05   \cite{doddington2004automatic}     & General                & 24    & 9k/1.5k             \\
% SemEval   \cite{hendrickx2010semeval}      & General                & 18    & 8k/2.7k           \\

% DDRel \cite{jia2021ddrel}& General                & 13    & 5k/1.2k           \\

% CMeIE \cite{zhang2022cblue}  & Medical                & 44    & 34k/8.7k            \\
% SCIERC \cite{luan2018multi}                & Scientific             & 7     & 3.2k/974           \\
% Wiki80 \cite{han2019opennre}               & Encyclopedic           & 80    & 12k/5.6k           \\
% NYT10   \cite{riedel2010modeling}          & General                & 53    & 455.4k/172.4k\\
NYT   \cite{riedel2010modeling}          & \textbf{\CheckmarkBold}     &   &    & 24    & 5.6k/5k & \textbf{\CheckmarkBold} $^3$ 
\\
WebNLG   \cite{gardent2017creating}          & \textbf{\CheckmarkBold}     &   &    & 171    & 5019/703 & \textbf{\CheckmarkBold} \\
WikiReading $\star$  \cite{hewlett2016wikireading}& \textbf{\CheckmarkBold}     &   &    & 884   & 14.85M/3.73M & \textbf{\CheckmarkBold} \\ 
% TACRED  \cite{zhang2017position}           & General                & 41    & 90.7k/15.5k    \\
WIKI-TIME \cite{yan2019relation}           & \textbf{\CheckmarkBold}     &   &    & 57   & 97.6k/40k &  \\
SciERC \cite{luan2018multi}           &     & Scientific    &         & 7     & 2,136/551  & \textbf{\CheckmarkBold}  \\
FOBIE \cite{kruiper2020layman}          &   & Scientific   &        & 3    & 1,238/300  & \textbf{\CheckmarkBold}  \\
DialogRE   \cite{yu2020dialogue}        & \textbf{\CheckmarkBold}     &   &        & 37    & 6k/1.9k  & \textbf{\CheckmarkBold}  \\
% SciREX $\star$ \cite{jain2020scirex}           &     & Scientific    &         & 21     & 306/132  & \textbf{\CheckmarkBold}  \\
FewRel 2.0   \cite{gao2019fewrel}          &  & Medical   &     & 100+25& 56k/14k   & \textbf{\CheckmarkBold} $^4$ 
\\
ChemProt   \cite{peng2019transfer}         &  & Biochemical  &          & 14    & 19.5k/16.9k  & \textbf{\CheckmarkBold}  \\
DDI   \cite{herrero2013ddi}       &   & Biochemical    &        & 5    & 25.3k/5.7k  & \textbf{\CheckmarkBold}  \\
DocRED $\star$ \cite{yao2019docred}           & \textbf{\CheckmarkBold}     &   &      & 96    & 4k/1k & \textbf{\CheckmarkBold}  \\
CUAD \cite{hendrycks2021cuad}    &   & Legal    &        & 25    & 10.48k/2.62k  & \textbf{\CheckmarkBold}  \\
FinRED \cite{sharma2022finred}   &  & Finance  &          & 29    & 5,699/1,068  & \textbf{\CheckmarkBold}  \\
% MNRE ~\cite{zheng2021multimodal}     & Multi-modal            & 23    & 12.2k/3.2k       \\
SMiLER   \cite{seganti2021multilingual}       & \textbf{\CheckmarkBold}  &  & \textbf{\CheckmarkBold}      & 36    & 733k/15k  &   \\
mLAMA   \cite{kassner2021multilingual}       & \textbf{\CheckmarkBold}  &  & \textbf{\CheckmarkBold}      & 5    & -   &  \\
ACE 2023  \cite{doddington2004automatic}       & \textbf{\CheckmarkBold}  &  & \textbf{\CheckmarkBold}      & 24    & 100k/50k  & \textbf{\CheckmarkBold}  \\
ACE 2024  \cite{doddington2004automatic}       & \textbf{\CheckmarkBold}  &  & \textbf{\CheckmarkBold}      & 24    & 300k/50k  & \textbf{\CheckmarkBold}  \\
\bottomrule
\end{tabular}
}}
% \vspace{-0.4cm}
\end{table}

\begin{table}[]
\caption{F1-scores (\%) of recent representative models on the NYT, WebNLG, and SciERC datasets. The models are categorized by their architectures, including Transformer-based models, and LSTMs marked with $\dagger$. }
\label{tab:performance_compare}
\small
\centering
\scalebox{0.9}{
\resizebox{\textwidth}{!}{%
\begin{tabular}{ p{2.4cm}  p{1cm}  p{4.1cm}  p{3cm} >{\raggedright\arraybackslash}p{2.5cm}}
\toprule
\textbf{Model} & \textbf{F1 (\%)} & \textbf{Text Representation} & \textbf{Context Encoding} & \textbf{Triplet Prediction}  \\ \midrule
\multicolumn{5}{c}{\textbf{\textit{NYT}}} \\ \midrule
\textbf{UniRel} \cite{ tang2022unirel} & 93.7 & Word-level, Position-enhanced & PLMs-based & Span-based \\ \midrule
\textbf{REBEL} \cite{ cabot2021rebel} & 93.4 & Word-level & PLMs-based & Seq2Seq \\ \midrule
\textbf{SPN} \cite{ sui2023joint} & 92.5 & Word-level, Position-enhanced & PLMs-based, Attention & Span-based \\ \midrule
\textbf{TDEER} \cite{ li2021tdeer} & 92.5 & Word-level, Character-level & CNN\&RNN, Attention & Sequence labeling \\ \midrule
\textbf{PFN} \cite{ yan2021partition} & 92.4 & Word-level, Position-enhanced & PLMs-based & Span-based \\ \midrule
\textbf{RIFRE} \cite{ zhao2021representation} & 92.0 & Word-level, Position-enhanced & CNN\&RNN, Attention & Sequence labeling \\ \midrule
\textbf{TPLinker} \cite{ wang2020tplinker} & 91.9 & Word-level, Position-enhanced & PLMs-based & Sequence labeling \\ \midrule
\textbf{RIN} \cite{ sun2020recurrent} & 87.8 & Word-level & RNN & Pipeline \\ \midrule
\multicolumn{5}{c}{\textbf{\textit{WebNLG}}} \\ \midrule
\textbf{UniRel} \cite{ tang2022unirel} & 94.7 & Word-level, Position-enhanced & PLMs-based & Span-based \\ \midrule
\textbf{PFN} \cite{ yan2021partition} & 93.6 & Word-level, Position-enhanced & PLMs-based & Span-based \\ \midrule
\textbf{SPN} \cite{ sui2023joint} & 93.4 & Word-level, Position-enhanced & PLMs-based, Attention & Span-based \\ \midrule
\textbf{TDEER} \cite{ li2021tdeer} & 93.1 & Word-level, Character-level & CNN\&RNN, Attention & Sequence labeling \\ \midrule
\textbf{RIFRE} \cite{ zhao2021representation} & 92.6 & Word-level, Position-enhanced & CNN\&RNN, Attention & Sequence labeling \\ \midrule
\textbf{TPLinker} \cite{ wang2020tplinker} & 91.9 & Word-level, Position-enhanced & PLMs-based & Sequence labeling \\ \midrule
\textbf{RIN} \cite{ sun2020recurrent} & 90.1 & Word-level & RNN & Pipeline \\ \midrule
\textbf{CGT} \cite{ ye2021contrastive} & 83.4 & Word-level, Position-enhanced & PLMs-based & Sequence labeling \\ \midrule
\multicolumn{5}{c}{\textbf{\textit{SciERC}}} \\ \midrule
\textbf{PL-Marker} \cite{ye2022packed} & 53.2 & Word-level, Position-enhanced & PLMs-based & Span-based \\ \midrule
\textbf{TriMF} \cite{ shen2021trigger} & 52.44 & Word-level, Position-enhanced & PLMs-based & Span-based \\ \midrule
\textbf{SpERT.PL} \cite{ santosh2021joint} & 51.25 & Word-level, Position-enhanced & PLMs-based, Attention & Span-based \\ \midrule
\textbf{SpERT} \cite{ eberts2020span} & 50.84 & Word-level, Position-enhanced & PLMs-based & Span-based \\ \midrule
\textbf{PURE} \cite{ zhong2021frustratingly} & 50.1 & Word-level & PLMs-based, Attention & Pipeline \\ \midrule
\textbf{DyGIE++} \cite{ wadden2019entity} & 48.4 & Word-level, Syntactic-enhanced & GNN, PLMs-based & Span-based \\ \midrule
\textbf{DyGIE} \cite{ luan2019general} & 41.6 & Word-level, Syntactic-enhanced & GNN, PLMs-based & Span-based \\ \midrule
\textbf{SciIE} \cite{ luan2018multi} & 39.3 & Word-level, Syntactic-enhanced & GNN & Span-based \\ \bottomrule
\end{tabular}
}}
\end{table}

\subsection{Datasets}
Annotated datasets are crucial for the development of RE methods.
We summarize the recently released and widely used benchmark datasets for RE in Table \ref{tab:datasets}\footnote{\url{https://paperswithcode.com/datasets}}\footnote{\url{https://nlpprogress.com/english/relationship_extraction.html}}\footnote{\url{https://thunlp.github.io/2/fewrel2_da.html}}, noting that the datasets listed are some representative examples and that many others also exist. 
Generally, these RE datasets can be roughly classified into four categories based on their data sources: (1) general corpora collected from news articles; (2) encyclopedic corpora collected from Wikipedia and Wikidata; (3) domain-specific corpora that contain scientific, finance, medical and biochemical articles; (4) multi-lingual corpora that include input texts in multiple languages; and (5) multi-modal corpora that contain textual relations with visual information.
Note that all datasets are manually annotated, except the NYT dataset \cite{riedel2010modeling}, which is created by a distant supervision approach using the knowledge base Freebase. 
Some follow-up works calibrated subsets of the NYT dataset to obtain more accurate annotation, like WIKI-TIME \cite{yan2019relation}.
Most existing datasets focus on sentence-level RE in general domain. Recently, some works have started to focus on the annotation and evaluation setups in more complex scenarios, including document-level \cite{yao2019docred,hewlett2016wikireading,jain2020scirex}, low-resource \cite{gao2019fewrel}, multi-modal \cite{zheng2021multimodal} and multi-lingual \cite{seganti2021multilingual} settings. 
To provide quantitative results, we include the corresponding leaderboard links in Table~\ref{tab:datasets}. Additionally, Table~\ref{tab:performance_compare} lists representative methods for the popular benchmark datasets, offering a clear benchmark for future research in this domain. 
Table~\ref{tab:performance_compare} highlights the effectiveness of Transformer-based models, which dominate the top performance across all datasets, reflecting their superior capability in handling complex language tasks.

\subsection{Evaluation Metrics}
The performance of supervised learning RE systems is typically measured by comparing the predicted labels to the corresponding ground-truth annotations. There are three main metrics \cite{taille2020let}: precision (P), recall (R), and F1 score. 
Specifically, P measures the proportion of correctly recognized results, while R assesses the proportion of all correctly recognized entities. The F1 score, being the harmonic mean of precision and recall, offers a balanced reflection of the system's performance.
For distant supervised relation extraction tasks, labels are generated automatically from external knowledge bases and may not be entirely accurate. As a result, metrics in supervised RE may not fully reflect the model's performance in real-world scenarios.
Therefore, Precision@$K$, the Precision-Recall Curve (PRC), and its Area Under Curve (AUC) \cite{zhu2020towards} are adopted as evaluation metrics for evaluating distantly supervised RE. 
Precision@$K$ measures the proportion of correctly identified relations among the top-$K$ predictions made by the system. 
For each instance, the RE model generates a ranked list of $K$ predictions, prioritizing them based on certain relevance criteria. The top $K$ predictions represent the subset of relations that the model deems most likely or relevant. However, evaluating Precision@$K$ requires manual effort, as researchers must annotate the top-$K$ output results of systems.
The PRC and AUC enable us to understand the precision-recall trade-off across various thresholds, comprehensively assessing the performance of distant-supervised RE models.  While the metrics presented are valuable and commonly used for evaluating RE systems, it is important to note that they are not exhaustive. Other metrics may also be relevant, depending on the specific characteristics of the task and the goals of the evaluation.

\section{Deep Learning Techniques for Relation Extraction}
\label{sec:dl_techniques_for_RE}
Recent advances in RE have largely been driven by DL techniques. In this section, we propose a new taxonomy to summarize DL-based RE approaches from three perspectives: text representation, context encoding, and triplet prediction. 
For each part, we present a comprehensive review of approaches in the literature. The subset of representative models illustrated in Figure~\ref{fig:architecture_in_our_taxonomy} serves as examples to illustrate this taxonomy.

\tikzstyle{mybox}=[
    rectangle,
    draw=hiddendraw,
    rounded corners,
    text opacity=1,
    minimum height=2em,
    minimum width=15em,
    inner sep=2pt,
    align=center,
    fill opacity=.5,
    ]
    
\tikzstyle{leaf}=[mybox,minimum height=1em,
fill=cyan!20, text width=20em,  text=black,align=left,font=\scriptsize,
inner xsep=2pt,
inner ysep=1pt,
]

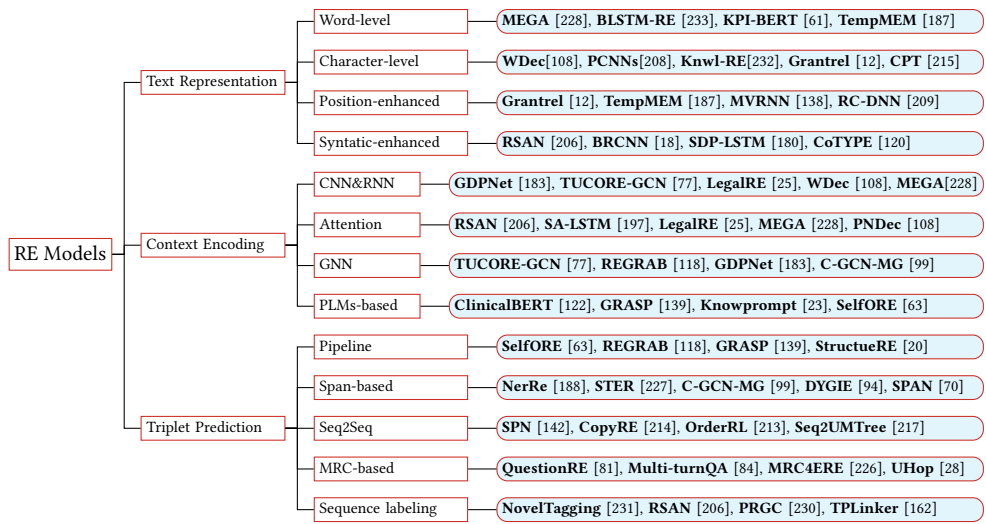
\begin{figure*}[tp]
\setlength{\abovecaptionskip}{5pt}   
\setlength{\belowcaptionskip}{5pt}
  \centering
  \scalebox{1}{
\resizebox{\textwidth}{!}{%
  \begin{forest}
    forked edges,
    for tree={
      grow=east,
      reversed=true,
      anchor=base west,
      parent anchor=east,
      child anchor=west,
      base=left,
      font=\small,
      rectangle,
      draw=hiddendraw,
      % rounded corners,
      align=left,
      minimum width=2.5em,
      % edge+={darkgray, line width=1pt},
      s sep=6pt,
      inner xsep=2pt,
      inner ysep=1pt,
      ver/.style={rotate=90, child anchor=north, parent anchor=south, anchor=center},
    },
    where level=1{text width=5.6em,font=\scriptsize}{},
    where level=2{text width=4em,font=\scriptsize}{},
    [RE Models
    [Text Representation
        [Word-level,text width=6em
           [\textbf{BLSTM-RE}~(2017)~\cite{zhou2017joint}{, }\textbf{TempMEM}~(2019)~\cite{yan2019relation}{, }\textbf{MEGA}~(2021)~\cite{zheng2021multimodal}{, }\textbf{KPI-BERT}~(2022)~\cite{hillebrand2022kpi}
            ,leaf,text width=29.2em]
        ]
        [Character-level,text width=6em
           [\textbf{Knwl-RE}~(2005)~\cite{zhou2005exploring}{, }\textbf{CPT}~(2006)~\cite{zhang2006composite}{, }\textbf{PCNNs}~(2015)~\cite{zeng2015distant}{, }\textbf{WDec}~(2020)~\cite{nayak2020effective}{, }\textbf{Grantrel}~(2021)~\cite{bian2021grantrel}
            ,leaf,text width=30.7em]
        ]
        [Position-enhanced,text width=6em
        [\textbf{MVRNN}~(2012)~\cite{socher2012semantic}{, }\textbf{RC-DNN}~(2014)~\cite{zeng2014relation}{, }\textbf{TempMEM}~(2019)~\cite{yan2019relation}{, }\textbf{Grantrel}~(2021)~\cite{bian2021grantrel}
            ,leaf,text width=29.2em]
        ]
        [Syntatic-enhanced,text width=6em
           [\textbf{SDP-LSTM}~(2015)~\cite{xu2015classifying}{, }\textbf{BRCNN}~(2016)~\cite{cai2016bidirectional}{, }\textbf{CoTYPE}~(2017)~\cite{ren2017cotype}{, }\textbf{RSAN}~(2021)~\cite{yuan2021relation}
            ,leaf,text width=29.2em]
        ]
      ]
      [Context Encoding
        [CNN\&RNN,text width=3.2em
           [\textbf{LegalRE}~(2020)~\cite{Chen2020JointEA}{, }\textbf{WDec}~(2020)~\cite{nayak2020effective}{, }\textbf{GDPNet}~(2021)~\cite{xue2021gdpnet}{, }\textbf{TUCORE-GCN}~(2021)~\cite{Lee2021GraphBN}{, }\textbf{MEGA}~(2021)~\cite{zheng2021multimodal}
            ,leaf,text width=33.1em]
        ]  
        [Attention,text width=3.2em
           [\textbf{SA-LSTM}~(2019)~\cite{yu2019beyond}{, }\textbf{LegalRE}~(2020)~\cite{Chen2020JointEA}{, }\textbf{PNDec}~(2020)~\cite{nayak2020effective}{, }\textbf{MEGA}~(2021)~\cite{zheng2021multimodal}{, }\textbf{RSAN}~(2021)~\cite{yuan2021relation}
            ,leaf,text width=32em]
        ]
        [GNN,text width=3.2em
                [\textbf{REGRAB}~(2020)~\cite{Qu2020FewshotRE}{, }\textbf{C-GCN-MG}~(2020)~\cite{mandya2020graph}{, }\textbf{TUCORE-GCN}~(2021)~\cite{Lee2021GraphBN}{, }\textbf{GDPNet}~(2021)~\cite{xue2021gdpnet}
                ,leaf,text width=32em]
        ]
        [PLMs-based,text width=3.4em
                [\textbf{SelfORE}~(2020)~\cite{hu2020selfore}{, }\textbf{ClinicalBERT}~(2021)~\cite{roy2021incorporating}{, }\textbf{GRASP}~(2022)~\cite{Son2022GRASPGM}{, }\textbf{Knowprompt}~(2022)~\cite{chen2022knowprompt}
                ,leaf,text width=32em]
        ]
      ]
      [Triplet Prediction
        [Pipeline,text width=6em
           [\textbf{SelfORE}~(2020)~\cite{hu2020selfore}{, }\textbf{REGRAB}~(2020)~\cite{Qu2020FewshotRE}{, }\textbf{StructueRE}~(2011)~\cite{chan2011exploiting}{, }\textbf{GRASP}~(2022)~\cite{Son2022GRASPGM}
            ,leaf,text width=29.2em]
        ]
        [Span-based,text width=6em
                [\textbf{DYGIE}~(2019)~\cite{luan2019general}{, }\textbf{C-GCN-MG}~(2020)~\cite{mandya2020graph}{, }\textbf{SPAN}~(2020)~\cite{Ji2020SpanbasedJE}{, }\textbf{NerRe}~(2022)~\cite{yan2022empirical}{, }\textbf{STER}~(2022)~\cite{zhao2022exploring} \quad\quad
                ,leaf,text width=30.5em]
        ]
        [Seq2Seq,text width=6em
                [\textbf{CopyRE}~(2018)~\cite{zeng2018extracting}{, }\textbf{OrderRL}~(2019)~\cite{zeng2019learning}{, }\textbf{SPN}~(2020)~\cite{sui2020joint}{, }\textbf{Seq2UMTree}~(2020)~\cite{zhang2020minimize}
                ,leaf,text width=29.2em]
        ]
        [MRC-based,text width=6em
                [\textbf{QuestionRE}~(2017)~\cite{levy2017zero}{, }\textbf{Multi-turnQA}~(2019)~\cite{li2019entity}{, }\textbf{UHop}~(2019)~\cite{chen2019uhop}{, }\textbf{MRC4ERE}~(2021)~\cite{zhao2021asking}
                ,leaf,text width=29.2em]
        ]
        [Sequence labeling,text width=6em
                [\textbf{NovelTagging}~(2017)~\cite{zheng2017joint}{, }\textbf{TPLinker}~(2020)~\cite{Wang2020TPLinkerSJ}{, }\textbf{RSAN}~(2021)~\cite{yuan2021relation}{, }\textbf{PRGC}~(2021)~\cite{zheng2021prgc}
                ,leaf,text width=29.2em]
            %\textbf{PRGC}~\cite{zheng2021prgc}{, }\\{, }\textbf{TER}~\cite{yu2019joint}
        ]
      ]
    ]
  \end{forest}
  }}
  \caption{The taxonomy of relation extraction models and the corresponding representative methods in each category are listed. }
  \label{fig:architecture_in_our_taxonomy}
  % \vspace{-0.5cm}
\end{figure*}

\subsection{Text Representation}
\label{subsec:input-representation}
For DL-based RE approaches, it is vital to learn powerful representations of the input data. 
Text representations encode each input token with a real-valued vector. Words that are similar in meaning are expected to be closer in vector space.
The ability of such distributed representations to capture syntactic and semantic properties of words affects the language modeling performance of DL-based RE approaches.  
We review and discuss the various types of text representation learning approaches used in previous RE works, including word-level, character-level, position-level, and syntactic-level representations.

\subsubsection{\textbf{Word-level Embeddings}}
Recent studies have demonstrated the importance of pre-trained word embeddings, which encode the meaning of input units into a real-valued continuous space. These word embeddings can be either fine-tuned or fixed during training. 

Non-contextualized word embeddings, such as Word2Vec \cite{mikolov2013distributed} and GloVe \cite{pennington2014glove}, are obtained by unsupervised algorithms, including continuous bag-of-word (CBOW) and continuous skip-gram models. 
These studies \cite{zeng2014relation, turian2010word} use high-dimensional distributed representations of words as input features for RE tasks, which encode the semantic information about entity words and help identify the relations among entities. For example,
% \cite{zheng2017joint} 
\citeauthor{zheng2017joint} \shortcite{zheng2017joint} proposed an end-to-end model to jointly extract entities and relations in a single model, constructing the word embeddings trained on the NYT corpus through the Word2Vec toolkit.
% \cite{zhou2017joint} 
\citeauthor{zhou2017joint} \shortcite{zhou2017joint} used the pre-trained 300-dimensional word vectors from Google in their proposed neural model for extracting relations.

Contextualized word embeddings of PLMs, such as BERT \cite{devlin2018bert} and ELMo \cite{peters-etal-2018-deep}, have demonstrated the importance of pre-trained word embeddings. These PLMs can be further fine-tuned during RE model training. A significant advantage is that the embeddings are contextualized by their surrounding text, meaning the same word can have different embedding depending on its contextual use.

\subsubsection{\textbf{Character-level Embeddings}}
To capture the sub-word level information, \textbf{character-level embeddings} \cite{nayak2020effective} are introduced to encode fine-grained information such as n-gram, prefix, and suffix features. 
Previous methods \cite{ zhang2006composite, zhou2005exploring, zeng2015distant}  explore the utilization of both internal and external contexts. In these studies, the sentence is partitioned into three segments based on the two entities of interest: the internal context encompassing characters within these entities, and the external context encompassing characters surrounding them.
Additionally, PLMs like BERT~\cite{devlin2018bert} inherently take subwords into account, further enriching the character-level representation and enabling the model to infer representations for unseen words. This characteristic is advantageous for handling out-of-vocabulary scenarios.

\subsubsection{\textbf{Position-level Embeddings}}
In addition, 
\citeauthor{yan2019relation} \shortcite{yan2019relation} 
proposed using \textbf{position-enhanced embeddings} for text representations in RE, and experimental results demonstrated that adding position information could sufficiently exploit the relative distance of the target entity pairs.
\citeauthor{yuan2019cross} \shortcite{yuan2019cross} encoded the position information in sentences, which can be formulated as follows: first, for a sentence, transform the word at position $i$ into a pre-trained word vector $v_i$ \cite{mikolov2013efficient}. Then, they calculate the relative distances to the target entities (i.e., $d_1$ and $d_2$) in the sentence and look up the position embedding table \cite{zeng2014relation} to find their position embeddings $p_{d_1}$ and $p_{d_2}$. Here, the position embedding table is randomly initialized and further updated during the processing of training. The word representation $w_i$ is represented by concatenating $v_i$ with $p_{d_1}$ and $p_{d_2}$. 
After repeating these steps, each sentence is transformed into a fixed-sized matrix $C=[w_i, w_2, \cdots, w_m]^T$, where  $m$ is the maximum length of the sentence in whole input data, and $w_i$ is a fixed-length vector. 
Sentences shorter than $m$ are padded with zero vectors.
\citeauthor{zeng2015distant} \shortcite{zeng2015distant, zeng2014relation} proposed using position embeddings for feature extraction in RE, and their results show that adding position information is superior to only using word information.
\citeauthor{zeng2014relation} \shortcite{zeng2014relation} exploited the position information to encode the relative distances to the target entity pairs.
\citeauthor{zhang2017position} \shortcite{zhang2017position} augmented the word representations with extra distributed representations of word position by combining the LSTM model with an entity position-aware attention.

\subsubsection{\textbf{Syntactic-level Embeddings}}
Moreover, another line of studies explored \textbf{syntactic-enhanced embeddings} to incorporate rich syntactic-related features into word embeddings, such as the shortest dependency path (SDP), Part-of-Speech (POS) tagging, WordNet hypernyms, and grammatical relations \cite{Chen2020JointEA,yuan2021relation}.
For example, 
\citeauthor{zeng2014relation} \shortcite{zeng2014relation} incorporated prior knowledge in texts, such as syntactic parsing and Part-of-Speech (POS) tagging, where the performance outperforms the baseline only using word-level representations.
\citeauthor{xu2015classifying} \shortcite{xu2015classifying} employed rich features in addition to word embeddings, including the SDP, POS tags, WordNet hypernyms, and grammatical relations, jointly integrating syntax and semantics. 
\citeauthor{xu2015semantic} \shortcite{xu2015semantic} constructed a comprehensive word representation by concatenating the word representation and the syntactical representation, which contains dependency labels and dependency edge directions. 
\citeauthor{cai2016bidirectional} \shortcite{cai2016bidirectional} and 
\citeauthor{nayak2020effective} \shortcite{nayak2020effective} 
focused on the syntactic structures in the input sentences, which were obtained by a dependency parser and provided complementary evidence for relationships. 
Their boosted performances demonstrate that adding additional information may lead to improvements in RE performance.
\citeauthor{ren2017cotype} \shortcite{ren2017cotype} 
proposed a domain-independent framework CoTYPE, which jointly embeds text features, type labels, entity and relation mentions. The entity and relation mentions with relevant candidate types are integrated into the model.

\subsubsection{Summary}
Word-level embeddings are commonly used as standalone representations, whereas other embeddings, such as character-level, position-level, and syntactic-level embeddings, are less frequently utilized on their own. While each type can represent specific aspects of textual information independently, hybrid embeddings combine multiple types of embeddings to capture a wider range of linguistic features, thereby enhancing the overall representation quality for RE tasks. However, blending features in hybrid embeddings can introduce complexity, potentially impacting the generality of neural RE models. The selection of external features depends on the specific application requirements.

\subsection{Context Encoding}
The word-level embeddings aim to extract lexical-level features from the given input data. 
Context encoding is designed to learn sentence-level features by capturing contextualized information and filtering out irrelevant information in the text representation. 
Context encoding can be implemented with any popular neural network architecture, such as CNNs, RNNs, attention-based neural networks, PLMs, and prompt tuning. These methods aim to retain almost all the information required to successfully predict the outputs.

\subsubsection{\textbf{Convolutional Neural Networks (CNNs) \& Recurrent Neural Networks (RNNs)}} 
CNNs \cite{zeng2014relation,zeng2015distant,shen2016attention,yuan2019cross} effectively learn local and position-invariant contextual representations. 
For example, \citeauthor{zeng2014relation} \shortcite{zeng2014relation} 
were among the first to use a convolutional deep neural network (CNN) for RE. It encodes the meaning of sentences not explicitly represented in the input representation.
\citeauthor{zeng2015distant} \shortcite{zeng2015distant} 
utilized a piecewise CNN model to scale hidden vectors for each word. The obtained feature vectors are then used to determine the relations through a feed-forward layer with a softmax function. 
\citeauthor{shen2016attention} \shortcite{shen2016attention} incorporated a combination of the CNN model and an attention network, which extracts the global features and attentive features in the sentence.
\citeauthor{yuan2019cross} \shortcite{yuan2019cross} adopted a piecewise-CNN (P-CNN) to consider the specific situation in RE, consisting of a convolutional layer and a Piecewise Max-pooling layer \cite{phi2019distant}.  Overall, CNNs are good at capturing local features within a sentence. However, CNNs may not capture long-distance dependencies efficiently, which is crucial in understanding complex sentence structures in RE tasks.

RNNs \cite{jat2018improving,nayak2020effective,zhang2017position,huang2015bidirectional,jat2018improving}, including long-short term memory (LSTM) and gated recurrent unit (GRU), have shown remarkable achievements in modeling sequential data. This property provides an excellent way to compose long context-dependent representations of sequence \cite{ma2016end}. 
\citeauthor{jat2018improving} \shortcite{jat2018improving} proposed a bidirectional gated recurrent unit (Bi-GRU) to extract the long-term dependency among the words in the input sequence.
The text representations encoded by bidirectional long-short term memory (Bi-LSTM) \cite{nayak2020effective} can efficiently incorporate the past and future text information \cite{huang2015bidirectional,jat2018improving}.
\citeauthor{zhang2017position} \shortcite{zhang2017position} introduced a position-aware attention mechanism over Bi-LSTM for the RE task, efficiently utilizing semantic similarity-based and position-based information.
Overall, RNNs are designed to handle sequential data, making them more suitable than CNNs for capturing long-range dependencies in text. They sequentially process words and can theoretically remember all previous information. However, in practice, RNNs often struggle with long sequences due to vanishing or exploding gradient problems, making it hard to capture very long-distance dependencies.

\subsubsection{\textbf{Attention-based Neural Networks}}
Attention-based neural networks \cite{yu2019beyond,shen2016attention} enhance the correlations between relation representations and text representations, highlighting important information for RE. 
Earlier studies \cite{shen2016attention, jat2018improving} incorporated word-level attention with sentence-level relation extraction. 
Recent works \cite{nayak2020effective,yu2019beyond,li2020self} combined attention networks with various models to capture multiple-grained entity and relation features. For example, 
\citeauthor{nayak2020effective} \shortcite{nayak2020effective} proposed a multi-focused attention model for RE, where dependency distance is incorporated to help identify the triplets in the input. 
The multi-factor attention helps focus on various pieces of evidence to determine the relationship.  
\citeauthor{yu2019beyond} \shortcite{yu2019beyond} introduced segment-level attention to select and model distributed representations of relational expressions.
\citeauthor{li2020self} \shortcite{li2020self} proposed a self-attention \cite{vaswani2017attention} enhanced model with entity-aware embeddings.
Overall, attention mechanisms allow models to focus on relevant parts of the text when predicting relations, effectively overcoming the limitations of CNNs and RNNs in handling long-range dependencies. They can capture complex sentence structures and relationships between entities regardless of their position in the text. However, these models can be computationally expensive and require a significant amount of data to train effectively.

\subsubsection{\textbf{Graph Neural Networks (GNNs)}}
GNNs \cite{welling2016semi,velivckovic2017graph,mandya2020graph, sahu2019inter} attempt to capture the non-linear structure of the input sequence by constructing semantic graphs, empowering the RE models with relational reasoning ability on graphs. 
GNN-based methods offer several key advantages, such as the ability to capture the global structure of the graph, and the ability to learn the representations of nodes and edges simultaneously.
Such graph-based models \cite{welling2016semi,velivckovic2017graph,zhang2018graph,guo2019attention} construct the non-linear structure of the input sequence via graphs, which provides a better way to represent the relationships between entities.
In particular, 
\citeauthor{zhang2018graph} \shortcite{zhang2018graph} utilized GCN and the syntactic dependency tree to construct the graph structure among the nodes. Then they built the adjacency matrix of the graph and included the edges from the shortest dependency path (SDP). 
\citeauthor{guo2019attention} \shortcite{guo2019attention} used a multi-head self-attention-based soft pruning strategy, which can identify the importance of edges in the graph. And some works
\cite{mandya2020graph, sahu2019inter, christopoulou2019connecting, nan2020reasoning} used the shortest dependency tree path to create the connections among nodes. 
Overall, GNNs can capture the interconnectedness of entities and relations in a way that is difficult for purely sequential models. However, constructing such graph structures requires additional preprocessing, and requires that relationships are easy to accurately represent in a graph.

\subsubsection{\textbf{Pre-trained Language Models (PLMs)}}
\label{sec:encoder_PLMs}
Recently, PLMs \cite{soares2019matching,Qu2020FewshotRE,zhao2021asking} have shown remarkable achievements in modeling RE problems by eliciting rich knowledge from large pre-trained models. PLMs usually trained on large-scale corpora, such as BERT \cite{devlin2018bert}, ELMo \cite{peters-etal-2018-deep}, RoBERTa \cite{liu2019roberta}, and SpanBERT \cite{joshi2020spanbert}, intrinsically incorporate auxiliary embeddings (e.g., position and segment embeddings).   
PLMs \cite{Qu2020FewshotRE,zhao2021asking} provide rich semantic knowledge to the RE task, where the fine-tuning process is performed on annotated task-specific data to adapt semantic information for RE. As PLMs are pre-trained on large amounts of text data, fine-tuning for relation extraction tasks \cite{du2018multi,wu2019enriching,soares2019matching} allows them better to understand the meaning and context of words and sentences. 
Particularly in scenarios with limited data availability, fine-tuning PLMs on the target task has proven to be an effective practice. Utilizing the information embedded in PLMs as the primary representation offers significant advantages in such cases.
One of the challenges is that downstream RE tasks fine-tuned on PLMs usually have different objective forms, leading to performance degradation. 
Prompt tuning \cite{chen2022knowprompt,Son2022GRASPGM} provides a new paradigm to stimulate the relation information of PLMs by bridging the format gap between the pre-training tasks and the downstream RE tasks. 
Recent works \cite{schick2020automatically,han2021ptr} show that prompt learning can effectively leverage the knowledge encoded in the PLMs, especially for few-shot RE tasks in Section~\ref{sec:FSRE}. 
Overall, PLMs offer substantial pre-trained knowledge that can capture intricate language patterns and dependencies. The main drawbacks of PLMs are their resource-intensive nature, requiring significant computational power for both training and inference, and their tendency to overfit on smaller or domain-specific datasets.

\subsubsection{Summary.}
Comparing the above encoders, there's a significant overlap in the application of these models, with many advanced systems combining their strengths. For example, earlier approaches combined RNNs or CNNs with attention mechanisms \cite{shen2016attention,zhang2017position} to capture both local features and global dependencies.
Integrating GNNs with attention mechanisms \cite{guo2019attention} allows for dynamic focus on different parts of the graph, enhancing the model's ability to capture complex relationships. PLMs \cite{devlin2018bert} inherently incorporate attention mechanisms like Transformers.  The choice of model often depends on the specific requirements of the task, including the nature of the data, the computational resources available, and the desired level of accuracy.

\begin{figure*}[t] 
	\centering 
	\includegraphics[width=0.95 \textwidth]{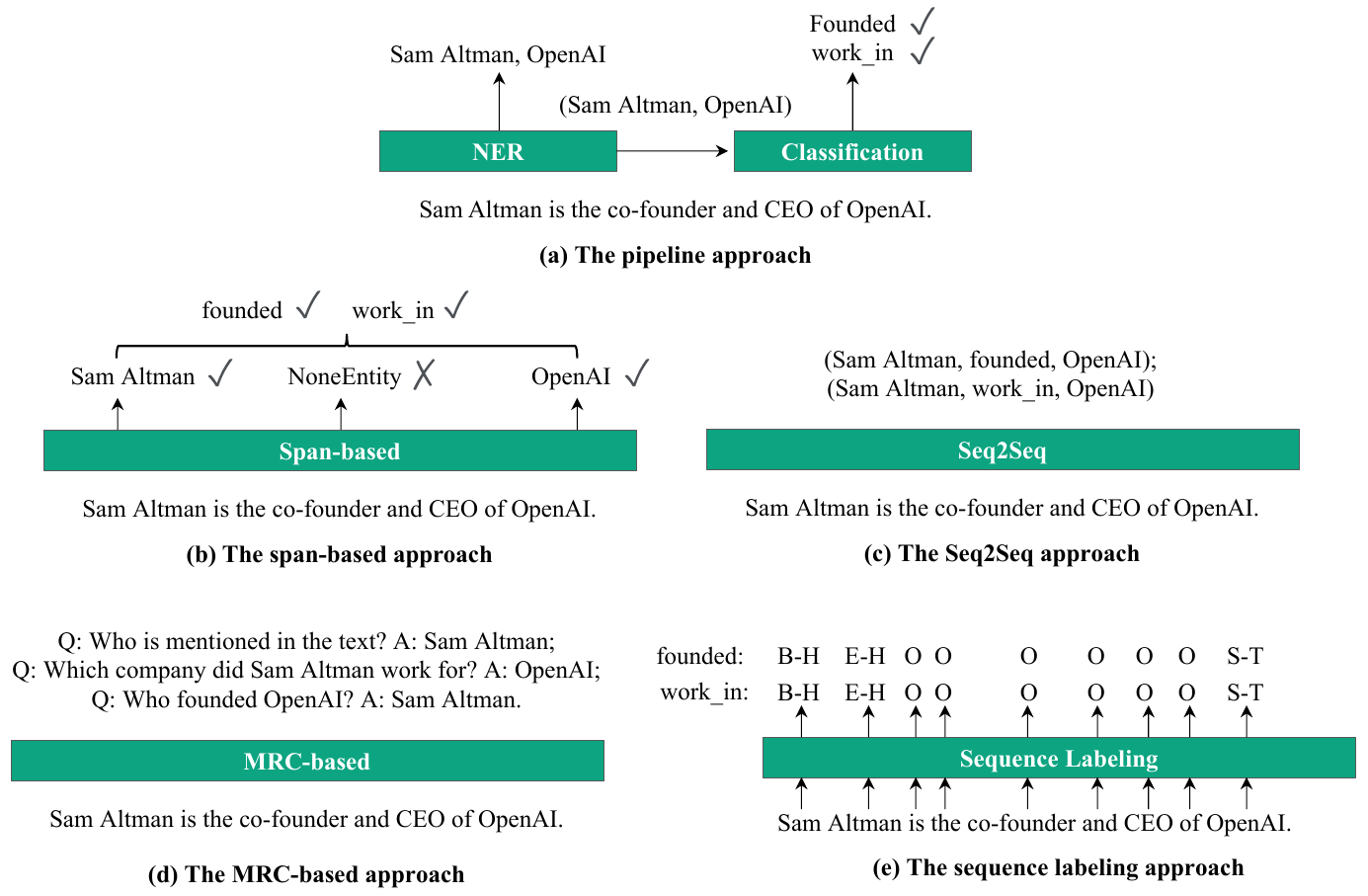} 
	\caption{Different relation extraction modeling paradigms: (a) is the pipeline-based approach; (b)-(e) are the joint approaches. The target triplet types in different RE models are shown. Each paradigm uses the input sentence ``Sam Altman is the co-founder and CEO of OpenAI.'' The target output triplets are \textit{<Sam Altman, founded, OpenAI>} and \textit{<Sam Altman, work\_in, OpenAI>}.}
	\label{fig:four_decoder_paradigm} 
\vspace{-0.4cm}
\end{figure*}

\subsection{Triplet Prediction}
The triplet prediction involves detecting the entity boundaries and classifying the relation types in the input sentence. 
Various modeling paradigms that have been proposed for decoding triplets in recent RE models. As illustrated in Figure \ref{fig:four_decoder_paradigm}, we group the existing triplet prediction paradigms into five categories, depending on the specific formulation of the relation extraction task. The corresponding types of target triplets for RE models are listed respectively.

\subsubsection{\textbf{The Pipeline-based (Classification) Approaches}}
The pipeline-based (classification) approaches separate the extraction of entities and relations \cite{miwa2016end,Son2022GRASPGM}. 
As illustrated in Figure \ref{fig:four_decoder_paradigm}(a), the pipeline approaches carry out entity recognition and relation classification sequentially \cite{miwa2016end,vashishth2018reside}. 
In the first stage, all candidate entities in the sentence are annotated manually or identified via NER models. Then, a classifier is used to determine the relation between every possible pair of identified entities. The ultimate goal is to accurately and consistently identify and extract all relevant relationships from the input text.
The pipeline approach assumes that entities are already identified, and models aim to identify the relationship (relations $R$ or $None$) between pairs of entities.

\begin{table}[]
\caption{Examples of No Entity Overlap (NEO), Single Entity Overlap (SEO) and Entity Pair Overlap (EPO) cases.}
\label{tab:overlap_examples}
\small
\begin{tabular}{ p{1cm} | p{5cm}| >{\raggedright\arraybackslash}p{6cm}}
\toprule
\multicolumn{1}{c}{\textbf{}}      & \multicolumn{1}{c}{\textbf{Text}}                                                             & \multicolumn{1}{c}{\textbf{Triplets}}                                                  \\ \midrule
\textbf{NEO} & The {[}United States{]} president {[}Donald   Trump{]} will visit {[}Beijing{]}, {[}China{]}. & (Donald Trump, President\_of, United   States) (China, Contains, Beijing)              \\ \midrule
\textbf{SEO} & The {[}United States{]} president {[}Donald   Trump{]} was born in {[}New York City{]}.       & (Donald Trump, President\_of, United   States) (Donald Trump, Born\_in, New York City) \\ \midrule
\textbf{EPO} & Martin went to {[}Tokyo{]} last week, which   is the capital of {[}Japan{]}.                  & (Japan, Contains, Tokyo)  (Japan, Capital,   Tokyo)                                  \\ \bottomrule
\end{tabular}
\end{table}

Different from the pipeline methods, joint-extraction approaches aim to find both entities and relations in a sentence by extracting valid relation triplets. These models face a challenge when extracting triplets from sentences with overlapping entities, which can be divided into three categories: (i) No Entity Overlap (NEO), where triplets do not share any entities; (ii) Single Entity Overlap (SEO), where at least two triplets share exactly one entity; and (iii) Entity Pair Overlap (EPO), where at least two triplets share some entities in the same or reverse order. 
A sentence can belong to both the SEO and EPO categories. 
As shown in Table~\ref{tab:overlap_examples} \cite{Yuan2020ARA}, the overlapping entities are marked in bold. The triplets in the second example (SEO class) share one single entity, Donald Trump. The triplets in the third example (EPO class) have overlapping entity pairs (Japan, Tokyo). 
Joint relation extraction aims to extract all relevant relation triplets present in a given sentence.

\subsubsection{\textbf{The Span-based Approaches}}
The span-based approaches~\cite{mandya2020graph,Dixit2019SpanLevelMF} process each sentence into spans and perform span classification to obtain predicted entities. Simultaneously, the detected entity pairs are regarded as candidate triplets for relation classification, as illustrated in Figure \ref{fig:four_decoder_paradigm}(b). Span-based approaches are shown with superior to previous pipeline-based methods \cite{yan2022empirical}. These methods \cite{Ji2020SpanbasedJE,ji2022span,Eberts2019SpanbasedJE} utilize pre-trained Transformer blocks to map word embeddings into BERT embeddings, calculate span and relation representations, perform classification and filtration tasks, and generate contextual semantic representations using multiple attention variants.
% \cite{zhao2022exploring}
\citeauthor{zhao2022exploring} \shortcite{zhao2022exploring} further define the privileged features in the RE task and propose a contrastive student-teacher learning framework to utilize the expert knowledge during training to enhance the performance of the model.
Some works \cite{luan2018multi,luan2019general,wadden2019entity} utilize dynamically constructed span graphs to achieve high performance on various tasks such as entity recognition and relation extraction. The most confident entity spans are selected and linked with confidence-weighted relation types and coreferences to construct the graphs, which iteratively optimize span representations.

\subsubsection{\textbf{The Seq2Seq-based Approaches}}
The Seq2Seq-based approaches~\cite{zeng2018extracting,zeng2019learning,zhang2020minimize,zeng2020copymtl,nayak2020effective} receive unstructured text as input and directly generate \textit{$\langle$head$\_$entity, relationship, tail$\_$entity$\rangle$} triplets as a sequential output. 
Formally, a source sentence $S = \{x_1, x_2, \cdots, x_n\}$ is represented as a sequence of words, where $x_i$ is the $i$-th word in $S$ and $n$ is the length of $S$. 
Based on the text representation (in Section \ref{subsec:input-representation}), the tag classifier predicts the relation types. The target sentence $T$ is represented as a sequence of words $T = \{t_1, t_2, \cdots, t_m\}$, where $t_j$ is the $j$-th word in $T$. Figure \ref{fig:four_decoder_paradigm}(c) shows the target triplet types in Seq2Seq-based models, which are able to tackle the overlapping relations and reduce the excessive computations. 
However, when dealing with tasks involving multiple triplets within a single sentence, the inherent linearization process of Seq2Seq models may pose challenges in processing extracts from multiple triplets with overlapping entities. To address these challenges, recent studies \cite{sui2020joint,zhang2020minimize} have focused on strategies to avoid the limitations imposed by the sequential nature of Seq2Seq models. 
For example, to deal with the overlapping problem, recent works \cite{Yuan2020ARA, zheng2021prgc,Wang2020TPLinkerSJ} design labeling strategies and perform the tagging process for multiple turns. These methods create specific sentence representations for each relation and then perform sequence labeling to extract the corresponding head and tail entities.

\subsubsection{\textbf{The MRC-based Approaches}}
MRC-based approaches \cite{li2019entity,zhao2021asking} treat the entity relation extraction task as a multi-turn QA task. For example, as shown in Figure \ref{fig:four_decoder_paradigm}(d), the relation type ``work\_in'' between ``Sam Altman'' and ``Open AI'' can be formulated as ``Question: Who is mentioned in the text? Answer: Sam Altman'' and ``Question: Which company did Sam Altman work for? Answer: Open AI''. 
Therefore, the extraction of entities and relations in a sentence can be transformed into the QA task of identifying answer spans from the context. This transformation allows the RE task to exploit well-developed machine reading comprehension (MRC) models \cite{li2019entity}, which extract text spans in passages given queries. For example,
\citeauthor{levy2017zero} \shortcite{levy2017zero} first formulated the RE task as a question answering (QA) task, where the relations are defined by natural-language question templates.
\citeauthor{li2019entity} \shortcite{li2019entity} 
 and \citeauthor{zhao2021asking} \shortcite{zhao2021asking} further transformed the RE task into a multi-turn QA task, providing a natural way to identify the entities and relations in a sentence. 
The RE process is thus converted into extracting information from textual passages by answering questions posed about the text. Additionally, approaches \cite{jiang2021new,chen2019uhop}  treat the task as a series of questions and answers, where each turn corresponds to a step in the extraction process. Overall, the key idea of MRC-based approaches is to formulate questions that prompt the model to identify relevant entities and relations within the text.

\subsubsection{\textbf{The Sequence Labeling Approaches}} 
Sequence labeling approaches solve RE task through shared parameters in an end-to-end manner, as illustrated in Figure \ref{fig:four_decoder_paradigm}(e). They perform joint RE by treating entity and relation types as well-designed tags \cite{zheng2017joint} and predict a single tag for each token. 
\citeauthor{yu2019joint} \shortcite{yu2019joint} tackled the joint RE extraction using an end-to-end sequence labeling framework based on functional decomposition. By breaking down the original task into smaller components, the learning process is simplified, resulting in improved overall performance, as presented by the empirical analysis in \cite{yu2019joint}.
To tackle the overlapping cases, 
some works \cite{yuan2021relation,hillebrand2022kpi,zheng2021prgc,Wang2020TPLinkerSJ,yu2019joint} perform sequence labeling in multi-turn by generating a specific tag sequence for each given relation.

\subsubsection{\textbf{Summary}} 
Previous RE surveys often overlook diverse decoding mechanisms. To bridge this gap, we provide a systematic survey of DL-based RE approaches focused on classifying the relations.
Overall, both pipeline-based and joint RE approaches exhibit their pros and cons.
The advantage of pipeline-based methods is that they are staged to detect named entities and classify relations, explicitly modeling the entity and relation information. However, the pipeline-based approaches assume that the entities are independent of relations, making them prone to accumulating errors and failing to capture the dependencies between entities and relations. 
In contrast, joint RE approaches are motivated by the fact that the entities and relations are closely related in real-world applications, thus avoiding error accumulation. 
Additionally, multiple relation triplets within an input text may share overlapping entities or relations.

\begin{table*}
 \caption{Overview of input and output for each RE task with examples.}
    \label{tab:examples}
	\centering
		{\renewcommand{\arraystretch}{1.0}
			\resizebox{1.0\columnwidth}{!}{
            \begin{tabular}{ p{1.2cm} | >{\raggedright\arraybackslash}p{1.4cm}| >{\raggedright\arraybackslash}p{2cm} | >{\raggedright\arraybackslash}p{4.5cm} | >{\raggedright\arraybackslash}p{1.2cm}| >{\raggedright\arraybackslash}p{3.5cm} }
            \toprule
                               & Task                     & Input             & Example Input*       & Output         & Example Output       \\
            \cline{1-6}
             \multirow{2}{1cm}{Low-resource RE}  & Distant Supervision RE   & A bag of sentences $S_b$   consisting of $b$ sentences and an entity pair $(e_1, e_2)$ presenting in all sentences. & Sentence-bag: \#1: Barack Obama was born in the United States. \#2: Barack Obama was the first African American to be elected to the president of the United States. \#3: Barack Obama served as the 44th president of the United States from 2009 to 2017. (\cite{shang2020noisy})    & Bag relation $r$ of the sentence-bag $S_b$ & president\_of     \\
             \cline{2-6}
            & Few-shot RE              & (Train on small support set $S$) Predict the relation $r$ for any given query instance $x$.  & In 2001, he also published the ”Khaki Shadows” that   recounted the military history of Pakistan during the cold war. (\cite{gao2019hybrid})    & Relation $r$  & Facet\_of \\
            \cline{1-6}
            Cross-sentence RE  & Document-level RE  &  Each sentence $d_i$ in a document $d$ & Lutsenko is a former minister of internal affairs. He occupied this post in the cabinets of Yulia Tymoshenko. The ministry of internal affairs is the Ukrainian police authority. (\cite{nan2020reasoning}) & The relation $r$ for each entity pair $(e_1, e_2)$ & (Lutsenko, manage, internal affairs) (Lutsenko, work\_with, Yulia Tymoshenko) (Yulia Tymoshenko, country\_of\_citizenship, Ukrainian)   \\
            \cline{2-6}
            & Dialogue RE & A dialogue $d = s_1:t_1, s_2:t_2, \cdots, s_m:t_m$ and entity piar $(e_1, e_2)$ &  S1: Hey Pheebs. | S2: Hey! | S1: Any sign of  your brother? | S2: No, but he’s   always late. | S1: I thought you only met him once? | S2: Yeah, I did. I think it sounds y’know big sistery, y’know, ‘Frank’s always late.’ | S1: Well relax,   he’ll be here.(\cite{yu2020dialogue})  & The relation $r$ of ($e_1$, $e_2$) based on $d$ &  (Frank, per:siblings,   S2) (S2, per:siblings,  Frank) (S2,   per:alternate\_names, Pheebs)      \\
            \cline{1-6}
            Domain-specific RE & RE in Biomedical Field   & A sentence $s$ is inserted with four makers.  &  Patient was given e11 ibuprofen e12 for high e21 fever e22. (\cite{roy2021incorporating}) (Note: e11, e12, e21, and e22 at the beginning and end of the target entities $(e_1,e_2)$.) & Relation $r$  & may\_treat \\
            \cline{2-6}
            & RE in Finance   Field    & A sentence $s$  &  MEXICO CITY —   State-owned oil company Pemex is reporting second quarter losses of \$US5.2   billion (\$A7.16 billion) due mainly lower petroleum prices. (\cite{sharma2022finred})   &  $(e_1, r, e_2)$ triplet set  & (Pemex, product\_or\_material\_produced,   petroleum) (Pemex, headquarters\_location, Mexico City) \\
            \cline{2-6}
            & RE in Legal Field  &  A sentence $s$ &  On August 19, 2014, Mr. Su sold methamphetamine to   Mr. Wang in Community A. (\cite{song2021})   & $(e_1, r, e_2)$ triplet set & (Mr. Su, traffic\_in, methamphetamine) (Mr. Su, sell\_drug\_to,   Mr. Wang) \\
            \cline{2-6}
            & RE in Scientific   Field & A sentence $s$  &  MORPA is a fully implemented parser method developed for a text-to-speech system. (\cite{luan2018multi})   &  $(e_1, r, e_2)$ triplet set  & (MORPA, Used\_for, text-to-speech system) (MORPA, Hyponym\_of,   parser)  \\
            \bottomrule  
            \end{tabular}
	}}
\end{table*}

\section{Challenging RE Problems and Solutions}
\label{sec:challenging_RE_problems_and_solutions}
Section \ref{sec:dl_techniques_for_RE} summarize the common practice for general RE problems. In this section, we review recent challenging RE problems and corresponding solutions. Table~\ref{tab:examples} shows the input and output for each challenging problem with examples.

\subsection{Low-resource Relation Extraction}
Supervised learning with DNNs requires a large-scale annotated training corpus which is difficult to
obtain in real-world applications, especially in 
low-resource settings. Recently, many efforts have been made to address low-resource RE. 

\subsubsection{\textbf{Distant Supervision Relation Extraction (DSRE)}}
DSRE aims to automatically leverage the facts in large-scale knowledge bases (KBs) to generate the annotated triplets as weak supervision. 
This technique can be traced back to the early work of \cite{mintz2009distant}, which proposed obtaining relationships for entity pairs aligned in KBs, such as Wikidata \cite{vrandecicWikidataFreeCollaborative2014}, DBpedia \cite{bizerDbpediaaCrystallizationPoint2009}, and Freebase \cite{bollackerFreebaseCollaborativelyCreated2008}. 
Despite the large amount of training data obtained through distant supervision, DSRE suffers from noisy label problems because individual sentences may give incorrect cues.
The noise present in this data mainly comes in two forms: (1) the obtained relations do not match the original meaning of the sentences, and (2) the relations and entities are missing due to incomplete knowledge bases. 

Existing DSRE studies mainly tackle the task at different granularities: 
(1) \textbf{Sentence-level}. These works \cite{du2018multi,yan2019relation,Bhartiya2021DiSReXAM} focus on finding accurate relational labels from the semantics of the input sentences. This approach is based on the strict assumption that if a pair of entities are found to share a relation in the KB, then any sentence containing that pair of entities is considered a positive instance of that relation. 
(2) \textbf{Bag-level}. This kind of approach is based on a slack assumption that at least one sentence in a ``bag’’ of sentences should express the relation.
There may exist several relations that can be chosen between specific entity pairs.
To mitigate the effects of noisy samples and make them more robust, 
\citeauthor{zeng2015distant} \shortcite{zeng2015distant} proposed a Piecewise Convolutional Neural Networks (PCNNs) model, which treats the distant supervised RE task as a multi-instance problem. The model avoids feature engineering and takes the uncertainty of instance labels into consideration. 
\citeauthor{yaghoobzadeh2017noise} \shortcite{yaghoobzadeh2017noise} proposed to address two types of noise from DS and pipeline input features, respectively. They introduced multi-instance multi-label learning algorithms to learn fine-grained entity typing and integrated entity typing into relation extraction to tackle the noise. To convert noisy labeling sentences into meaningful training data, 
\citeauthor{shang2020noisy} \shortcite{shang2020noisy} proposed an unsupervised deep clustering to produce new high-confidence relation labels for noisy sentences.
\citeauthor{yu2020tohre} \shortcite{yu2020tohre} formulated the DSRE as a hierarchical classification task and constructed the hierarchical bag representation to extract relations in a top-down manner.

Additionally, some works \cite{yuan2019cross,wu2019improving} that consider both sentence-level and bag-level information simultaneously, explore explicit cross-level interactions to further improve the performance of DSRE.
For example, 
\citeauthor{yuan2019distant} \shortcite{yuan2019distant} first used a linear attenuation simulation to reflect words' importance, then proposed a non-IID relevance embedding to capture the mutual information of instances in the bag.  
\citeauthor{ye2019distant} \shortcite{ye2019distant} proposed intra-bag and inter-bag attention models to address the noisy bag problem in a multi-instance distant supervision setting. 
\citeauthor{yuan2019cross} \shortcite{yuan2019cross} first employed sentence-level selective attention to reduce the effect of noise, then adopted cross-bag selective attention to capture the entity pairs with higher quality. 
\citeauthor{gou2020dynamic} \shortcite{gou2020dynamic} applied a dynamic parameter-enhanced network to DSRE, dynamically determining the sentence information to alleviate the style shift problem for predicting the long-tail relations.
\citeauthor{zhao2020cfsre} \shortcite{zhao2020cfsre} proposed context-aware based on frame semantics to combine the semantic knowledge within a hierarchical neural network to alleviate the noisy labels in DSRE.
\citeauthor{Dai2022CrossstitchingTA}\shortcite{Dai2022CrossstitchingTA} employed a cross-stitch mechanism to capture the interaction between the text encoder and knowledge graph (KG) encoder, allowing the model to share the information thoroughly.
\citeauthor{shang2023learning} \shortcite{shang2023learning} constructed a force-directed graph and introduced the attractive force to learn the correlation and mutual exclusion between different relations.

\textbf{Summary}. 
The development of DSRE has been characterized by continuous efforts to improve the accuracy and robustness of relation extraction models in the face of noisy and incomplete data. Researchers have explored various methodologies and advanced models to enhance DSRE's performance in extracting relational information from large-scale knowledge bases.

\subsubsection{\textbf{Few-shot Relation Extraction (FSRE)}}
\label{sec:FSRE}
FSRE aims to predict the relationship between two entities in a sentence by training with a few labeled instances for each relation. 
In realistic scenarios, only common relationships can obtain enough labeled examples, while most other relationships have very limited relational facts. 
FSRE has the potential to handle ``long-tail'' relations that have limited relational facts. 
In this section, we systematically present advanced FSRE approaches by categorizing them into two groups: 1) Metric Learning and 2) Knowledge-enhanced Learning. Additionally, we will discuss the most recent prompt-based methods with PLMs for few-shot RE tasks in Section \ref{sec:PLMs}. 

\textbf{1) Metric Learning}. 
One popular approach for FSRE is metric learning \cite{liu2022learn,Qu2020FewshotRE}, where the term ``metric'' refers to the distance function used to measure the similarity or dissimilarity between samples in the embedding space.  The model is optimized by minimizing the distances between query samples and their corresponding class prototypes, thereby improving its ability to assign new instances to the nearest relation class prototype. 
A relevant approach is learning prototypes of relations from the contextual information for capturing the semantics of relations, which significantly improves accuracy \cite{ding2020prototypical}.
Some approaches \cite{zhao2024few,gao2019hybrid,yu2020bridging,yu2022dependency} also introduce external information to compensate for the limited information in FSRE. For example, 
\citeauthor{Qu2020FewshotRE} \shortcite{Qu2020FewshotRE} 
proposed a global relation graph with text descriptions of entities and relations collected from Wikidata. 
\citeauthor{gao2019hybrid} \shortcite{gao2019hybrid} proposed a hybrid attention-based prototypical network to tackle the noisy problem in FSRE. They designed instance-level and feature-level attention to highlight important instances and features.
\citeauthor{yu2020bridging} \shortcite{yu2020bridging} proposed a multi-prototype embedding network to jointly extract relation triples. The prototype representations learned by specifical prototype-aware regularization can inject the implicit correlations between entities and relations. 
MapRE \cite{dong2021mapre} considered both label-agnostic and label-aware semantic mapping information for few-shot relation extraction. 
HCRP \cite{han2021exploring} learned relation label information by contrastive learning and allowed the model to adaptively learn to focus on hard work.
To endow a new model with the ability to optimize rapidly, REGRAB \cite{Qu2020FewshotRE} proposes a Bayesian meta-learning method by incorporating an external global relation graph. 
Overall, these approaches leverage metric learning strategies to learn a distance metric or similarity function, which can not only effectively distinguish between different types of relations but also facilitate better generalization from a limited number of labeled data for certain relations.

\textbf{2) Knowledge-enhanced Learning}. 
Many FSRE works also employ external knowledge to enrich the auxiliary semantic information. According to the data structure, external knowledge can be divided into (1) unstructured text spans,  
including the descriptions of entity and relation, 
and (2) a structured knowledge graph (KG).
For unstructured text span, TD-proto \cite{yang2020enhance} proposes a collaborative attention module to enhance the prototypical network with entity and relation descriptions. ConceptFERE \cite{yang2021entity} model introduces the inherent concepts of entities to provide appropriate clues for relation classification, bridging the gap between the representations of relation types and text.
\citeauthor{Wang2022DRKDR} \shortcite{Wang2022DRKDR} proposed a discriminative rule-based knowledge method where a logic-aware inference module is adopted to avoid the adverse effect of text features.
In comparison, some approaches explore the abundant KG information. 
\citeauthor{liu2020k} \shortcite{liu2020k} proposed to inject triples in KG into texts, which transforms the sentences into knowledge-enhanced sentence trees. 
\citeauthor{Roy2021IncorporatingMK} \shortcite{Roy2021IncorporatingMK} incorporated entity-level KG into pre-trained BERT for clinical RE, integrating the medical knowledge by several techniques.
\citeauthor{sainz2021label} \shortcite{sainz2021label} reformulated the RE task as an entailment task with hand-made verbalizations of relation labels, which helps generalize to the unseen label.  
Knowledge-enhanced RE methods leverage external knowledge sources to deepen the understanding of entity relations. However, this integration presents challenges, including managing noise and ensuring effective incorporation. Errors may arise from mapping concepts from external sources to the RE task, mainly due to semantic mismatches. Additionally, the complexity of filtering relevant information while discarding noise impacts efficiency and adds to the difficulty of the algorithm design.

\textbf{Summary}. 
The development of FSRE has been driven by the need to extract relational information from text when only a limited number of labeled instances are available for training. FSRE approaches leverage metric learning and knowledge-enhanced learning strategies to address the challenge of limited labeled data for relation extraction. These methods advance the field by effectively generalizing from a small number of labeled instances and enriching semantic understanding with external knowledge sources.

\subsection{Cross-sentence Relation Extraction}
While most existing works have focused on extracting relational facts from individual sentences, many relational facts are expressed across multiple sentences within a long context. As a result, many studies have shifted research attention from sentence-level to cross-sentence. 
Cross-sentence RE aims to overcome the inherent limitations of sentence-level approaches and identify all relations mentioned across multiple sentences.
Generally, there are two main research lines in cross-sentence RE: document-level relation extraction and dialogue relation extraction.

\subsubsection{\textbf{Document-level Relation Extraction}}
Document-level RE (DocRE) aims to extract the triplets mentioned in a document. Although sentence-level RE approaches have achieved impressive results \cite{li2019entity,nan2020reasoning}, they still fail to handle the document-level RE, as the documents contain richer information and more complex structures than sentences. 
Unlike sentence-level RE, which aims to classify the relations of one or several entity pairs, document-level RE requires the RE model to identify and focus on the relevant context within the document for a specific entity pair.    
Additionally, one entity pair can appear multiple times within a document, with each appearance having a distinct relation, making document-level RE more challenging than sentence-level RE.
For instance, as shown in Figure \ref{fig:intro_example},  the relation ``\textit{worked\_on}'' between ``\textit{Sam Altman}'' and ``\textit{ChatGPT}'' can only be founded in the long context of the document instead of the single sentence. Other sentences between these two sentences may contain irrelevant information. 
This requires the RE models to be capable of capturing long-distance dependency in relational information.

Recent RE methods \cite{Han2020MoreDM,yuan2021relation}  effectively capture complex interactions requiring reasoning over long-distance entities across multiple sentences. 
There are mainly two methods to infer relations from multiple sentences: 

1) \textbf{The Graph-based Approaches}. These works \cite{christopoulou2019connecting,nan2020reasoning} construct document graphs with attention or dependency structures, bridging the entities spreading far apart in the document.
Relational inference information is gathered from the graph neural networks.
These methods perform multi-hop reasoning in the overall graph structure to obtain meaningful entity representations. 
For example, 
\citeauthor{Nan2020ReasoningWL} \shortcite{Nan2020ReasoningWL} automatically constructed a document-level graph to empower relational reasoning across sentences. To enable the model with multi-hop reasoning, they proposed a refinement strategy to incrementally aggregate relevant information.
\citeauthor{Christopoulou2019ConnectingTD} \shortcite{Christopoulou2019ConnectingTD} proposed an edge-oriented graph neural model to construct a document-level graph using multiple nodes and edges.
\citeauthor{Tran2020TheDH} \shortcite{Tran2020TheDH} extended the edge-oriented model by incorporating explicit relation classification-related node representations.
\citeauthor{Li2020GraphED} \shortcite{Li2020GraphED} proposed a graph-enhanced dual attention network to characterize the complex interactions among potential relation instances. 
\citeauthor{zeng2020double} \shortcite{zeng2020double} designed a graph aggregation-and-inference network featuring a double graph. There is a heterogeneous mention-level graph to capture the interactions among different mentions and an entity-level graph to infer relations between entities. 
To cover more logical reasoning chains, 
\citeauthor{zeng2021sire} \shortcite{zeng2021sire} 
developed a logical reasoning module to represent intra- and inter-sentential relations.

\textbf{2) The Path-based (non-graph) Approaches}. These works \cite{jain2020scirex,Chen2020JointEA} attempt to enrich the local contextual information surrounding the target entity pair. 
They extract paths connected to the given entities to retain semantic information for predicting the relations. For example, 
\citeauthor{Yao2019DocREDAL} \shortcite{Yao2019DocREDAL} utilized context-aware LSTM to encode sentences and infer relations in the document. 
\citeauthor{Xu2021EntitySW} \shortcite{Xu2021EntitySW} 
formulated the distinctive dependencies by incorporating the structural dependencies based on the self-attention mechanism.
Moreover, some works explore synthesizing implicit long-distance information modeled by transformer-based methods and multi-scale neural architectures~\cite{jia2019document}. 
\citeauthor{xu2021document} \shortcite{xu2021document} proposed an encoder-classifier-reconstructor model for DocRE, where the reconstructor is used to reconstruct the path dependencies from the graph representation.  
To tackle the multi-label and multi-entity problem in DocRE, 
\citeauthor{zhou2021document} \shortcite{zhou2021document} deployed adaptive thresholding and localized context pooling to transfer attention from PLMs to decide the context relevant to the relation.
\citeauthor{Wang2022AUP} \shortcite{Wang2022AUP} constructed a unified positive-unlabeled learning method to tackle the incomplete labeling problem in DocRE.
\citeauthor{Chen2022IterativeDI} \shortcite{Chen2022IterativeDI} introduced an iterative extraction for document-level RE and proposed an imitation learning to cast the extraction problem as a Markov decision process.
\citeauthor{Guo2022DOREDO} \shortcite{Guo2022DOREDO} discovered that the inadequate training paradigm leads to underwhelming performance instead of the model capacities. Therefore, they propose a generative framework for document-level RE which generates a symbolic sequence from a relation matrix to help model learning.

\textbf{Summary}. 
Graph-based RE approaches construct document-level graphs and utilize graph structures to model the complicated relationships among entities. Such graph representations efficiently capture local and global contextual information, thereby facilitating the discovery of implicit relations. However, the construction and maintenance of graph structures can be computationally demanding, and the quality of the underlying graph representation plays a crucial role in determining the effectiveness of graph-based models. Conversely, path-based (non-graph-based) approaches concentrate on the sequential context and semantic patterns within candidate entity pairs, employing syntactic dependency structures or pre-trained models to connect target entities directly or through contextual tokens. These approaches generally exhibit greater computational efficiency and adaptability for various relation extraction tasks. 
However, path-based approaches may be less proficient than graph-based approaches at capturing global relational information. This limitation arises due to the sequential nature of path-based approaches, which may struggle to comprehensively capture relationships spanning distant parts of the document. 
Overall, both graph-based and path-based approaches have shown effectiveness in document-level RE tasks, with the potential for further enhancement through document understanding and the integration of multi-hop reasoning capabilities for inferring complex relationships.

\subsubsection{\textbf{Dialogue Relation Extraction}}
In addition to extracting semantic relations from sentences and documents, recent RE research also explores dialogue scenarios. The relation triplets in dialogue usually have low information density and do not appear simultaneously. This suggests that dialogue RE should be aware of the multiple speakers and arguments within a conversation. 
The represented dialogue RE approaches can be divided into two categories: 
(1) Fine-tuning PLMs with specific dialogue RE objectives. To capture the diverse relational information between arguments in the dialogue, some strategies are applied to build an RE model that obtains the contextualized turn representations  \cite{Lee2021GraphBN}, such as constructing the latent multi-view graph and heterogeneous dialogue graph. 
\citeauthor{Cai2019GraphTF} \shortcite{Cai2019GraphTF} proposed a graph transformer to explicitly encode relations and enable direct communication between distant node pairs.  
\citeauthor{Yao2020HeterogeneousGT} \shortcite{Yao2020HeterogeneousGT} proposed a heterogeneous graph transformer to model the different relations among individual subgraphs, including direct, indirect, and possible relations between nodes. 
(2) Prompt-based approaches.
These utilize prompting exemplars constructed with trainable words to incorporate potential relational knowledge. For example, 
\citeauthor{chen2022knowprompt} \shortcite{chen2022knowprompt} 
injected knowledge among the relation labels into prompting. 
\citeauthor{Son2022GRASPGM} \shortcite{Son2022GRASPGM} 
proposed an argument-aware prompting strategy to capture the relational clues. 

Furthermore, due to the properties of low information density and high personal pronoun frequency \cite{Lee2021GraphBN} in dialogue, more research efforts are needed to capture such sparse semantics among multiple speakers. 
\citeauthor{albalak2022d} \shortcite{albalak2022d} proposed a model-agnostic framework D-REX that focuses on dialogue RE and the explainability of methods. D-REX frames RE as a reranking task and incorporates relation- and entity-specific explanations in the intermediate steps. 
\citeauthor{yu2020dialogue} \shortcite{yu2020dialogue} defined the trigger words in dialogue RE which indicates the existence of a given relation. They showed such manually annotated text spans play a critical role in cross-sentence RE.
\citeauthor{Xue2020AnES} \shortcite{Xue2020AnES} took a novel input format and utilized a BERT-based model to capture the interrelations among entity pairs.
Additionally, some studies utilize token-graph models to track the speaker-related information for cross-sentence RE in dialogues.
\citeauthor{Chen2020DialogueRE} \shortcite{Chen2020DialogueRE} deployed a token graph attention network.
\citeauthor{Xue2020GDPNetRL} \shortcite{Xue2020GDPNetRL} proposed capturing relationships by generating a latent multi-view graph and selecting critical words for RE.
\citeauthor{qiu2021socaog} \shortcite{qiu2021socaog} proposed an $\alpha-\beta-\gamma$ strategy, an incremental parsing strategy for dynamic inference upon any incoming sentence, to infer social relations in dialogues. This strategy models the social network as a graph to ensure the consistency of relations.

\textbf{Summary.} 
Dialogues typically encompass complex discourse structures, implicit relations behind conversations, and dynamic interactions between speakers. Existing studies predominantly focus on extracting relations among various speakers or individuals mentioned in conversations, leveraging contextualized representations to capture intricate relationships within dialogues. However, due to the properties of low information density and high personal pronoun frequency in dialogue, more research efforts are necessary to effectively capture relational clues within dialogues, particularly in capturing sparse semantics among multiple speakers.

\subsection{Domain-specific Relation Extraction}
In real-world scenarios, RE approaches are typically applied to different specific domains. 
However, general-purpose RE models, when directly applied to domain-specific data, can yield unsatisfactory results due to the shift in word distribution from general domain data to domain-specific data.
Therefore, it is necessary to explore how to endow RE models with the ability to adapt to domain-specific corpora. 
Although RE studies have been thriving for a few decades, few researchers have reviewed domain-specific fields so far.
In this section, we discuss recent RE methods tailored to different specific domains, including biomedical \cite{guChemicalinducedDiseaseRelation2016, liNeuralJointModel2017, choiExtractionProteinProtein2018}, financial \cite{velaConceptRelationExtraction2009}, and legal \cite{andrewAutomaticExtractionEntities2018c}, and scientific domains.

\subsubsection{\textbf{RE in the Biomedical Field}}
In the biomedical field, the RE models aim to automatically extract relations between biomedical entities (proteins, genes, diseases, etc.) from a rich source of biomedical texts \cite{roy2021incorporating,wang2020learning}. 
BioBERT~\cite{lee2020biobert} is a representative model of PLMs that inject biomedical information.
Most works on biomedical RE focus on one type of relation, which can be categorized into several types according to biomedical relation types. These types include drug-drug interaction RE~\cite{asada2022integrating}, disease-protein RE~\cite{xu2019translating}, chemical-protein RE~\cite{weber2021humboldt}, and protein-protein interaction RE~\cite{ahmed2019identifying}.  
For example, 
\citeauthor{asada2022integrating} \shortcite{asada2022integrating} utilized heterogeneous domain information for drug-drug interaction RE, combining drug description and molecular structure information.
\citeauthor{weber2021humboldt} \shortcite{weber2021humboldt} defined the Humboldt contribution task as an RE problem, where the chemical-protein relations are modeled with PLMs by incorporating entity descriptions. 
\citeauthor{ahmed2019identifying} \shortcite{ahmed2019identifying} designed a tree LSTM model with structured attention architecture for identifying protein-protein interaction relationships. 
\citeauthor{zhao2021biomedical} \shortcite{zhao2021biomedical} explored modeling the global dependency relation of sentences by self-attention mechanism and graph convolutional networks. 
\citeauthor{Haq2021DeeperCD} \shortcite{Haq2021DeeperCD} introduced accuracy-optimized and a speed-optimized architecture. 
The systems understand different aspects of clinical documents, thereby enhancing the accuracy of extracting entity pairs and clinical relations, including extracting and correlating dates to generate a timeline of a patient’s data, as well as parsing and comprehending trial results for analysis.

Recently, with the success of PLMs, several Transformer-based approaches have been widely explored for biomedical RE.
\citeauthor{Wei2019RelationEF} \shortcite{Wei2019RelationEF} first explored implementing the BERT model for clinical RE tasks, where the unstructured clinical data is typically documented by specific professionals.
\citeauthor{thillaisundaram2019biomedical} \shortcite{thillaisundaram2019biomedical} 
proposed extracting biomedical triplets with an extended BERT model, which encoded gene-disease pairs and their textual context to predict the ``function change'' relation. 
\citeauthor{yadav2020relation} \shortcite{yadav2020relation} proposed a multi-task learning framework for relation extraction in biomedical and clinical domains, modeling the RE task with three subtasks to better utilize the shared representation.
\citeauthor{Kanjirangat2021EnhancingBR} \shortcite{Kanjirangat2021EnhancingBR} proposed a distantly supervised biomedical RE method using the shortest dependency path for selecting representative samples.
Moreover, 
\citeauthor{Sarrouti2022ComparingEA} \shortcite{Sarrouti2022ComparingEA} did an empirical study on encoder-only and encoder-decoder transformers over ten biomedical RE datasets. These comparisons also included the four major biomedical subtasks, i.e., chemical-protein RE, disease-protein RE, drug-drug RE, and protein-protein RE. They further explored multi-task fine-tuning to examine correlations among these subtasks. 

In addition, similar to the idea discussed in the knowledge-enhanced methods subsection~\ref{sec:FSRE}, in the biomedical field, KGs play a significant role in enriching manually annotated information \cite{dai2019distantly,milosevic2022relationship}. They offer substantial potential for leveraging external knowledge sources to enhance our understanding of entity relations and extract biomedical relationships.
Besides, there is an increasing demand to extract n-ary relations \cite{jia2019document} from multiple documents, where $n > 2$. 
It is essential to extract relations between more than a pair of entities in the biomedical field. For example, detecting the relationship between a drug, a cancer patient, and a specific gene mutation is crucial for determining whether a drug is relevant for treating cancer patients with a certain mutation in a given gene. 
\citeauthor{lee2020cross} \shortcite{lee2020cross} proposed cross-sentence N-ary RE by utilizing entity linking and discourse relations, respectively. 
\citeauthor{tiktinsky2022dataset} \shortcite{tiktinsky2022dataset} proposed an N-ary drug combination RE dataset to assist professionals in identifying beneficial drug combinations. 
They also proposed a baseline model to predict if a subset of drugs used together in combination therapy is effective.

\subsubsection{\textbf{RE in the Finance Field}}
In the financial domain, the RE systems focus on identifying specific relations within financial texts, such as automatically extracting and linking key performance indicators (KPIs) from financial documents \cite{hillebrand2022kpi}. For example,
\citeauthor{deusser2022kpi} \shortcite{deusser2022kpi} explored extracting KPIs from financial documents where a word-level weighting scheme models the inherently fuzzy borders of the entity pairs and the corresponding relations.
\citeauthor{Wu2020CreatingAL} \shortcite{Wu2020CreatingAL} focused on Chinese financial entity recognition and relation extraction and proposed a mixed pattern with POS tagging is proposed to generate the quadruples (entity1, entity2, relation, text) from the unstructured finical text.
\citeauthor{Jabbari2020AFC} \shortcite{Jabbari2020AFC} presented a domain-specific ontology for financial entities and relations in French news and created a corpus to build a knowledge base of financial relations.
\citeauthor{sharma2022finred} \shortcite{sharma2022finred} 
released the first financial RE dataset and demonstrated that the RE models trained on general domain might be ineffective in understanding financial relations in texts due to the discrepancies in the set of relations. 
The research value of financial relation extraction is to make full use of financial information and help investors make better investment decisions.

\subsubsection{\textbf{RE in the Legal Field}}
In the legal domain, RE systems aim to extract the legal relationships between entities contained in judicial documents, such as the relationship between a person and a company.
To automatically identify entities and relations in legal documents, 
\citeauthor{Andrew2018AutomaticEO} \shortcite{Andrew2018AutomaticEO} explored combining statistical and rule-based techniques without labeled data. 
\citeauthor{hendrycks2021cuad} \shortcite{hendrycks2021cuad} 
created the first legal dataset for contract reviews.
Previous studies focused on modeling implicit relations within legal documents, such as criminal relations in judgment documents \cite{Chen2020JointEA} and clause relations in contracts \cite{xu2022conreader}.
\citeauthor{Thomas2021SemisupervisedKP} \shortcite{Thomas2021SemisupervisedKP} proposed a semi-supervised pattern-based learning method to extract relational facts from the judicial text. This work combines bootstrapping and OBIE techniques to expedite the extraction of judicial facts.
\citeauthor{Wang2021CrossDomainCE} \shortcite{Wang2021CrossDomainCE} focused on cross-domain contract element extraction and proposed a Bi-FLEET model, which incorporates a clause-element relation encoder with a bi-directional feedback scheme. A multi-task framework is applied to capture interactions between contract element extraction and clause classification.
\citeauthor{Xu2022ConReaderEI} \shortcite{Xu2022ConReaderEI} proposed a ConReader framework for a better contract understanding, which explores the long-range context relation, term-definition relation, and similar clause relation in the contract clause extraction. 

\begin{figure}[t] 
	\centering 
	\includegraphics[width=0.8 \textwidth]{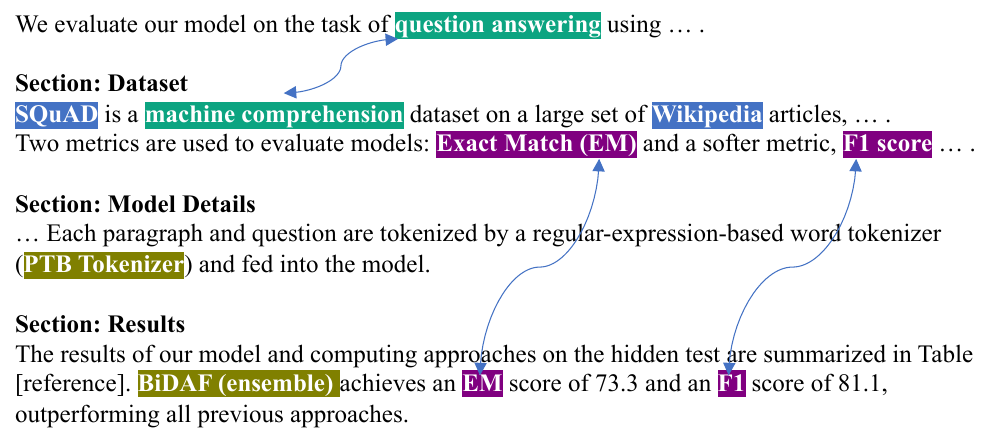} 
	\caption{An example in \cite{jain2020scirex} for the  document-level $N$-ary relation (Dataset: SQuaD, Task: Machine Comprehension, Method: BiDAF (ensemble), Metric: EM/F1). }
	\label{fig:n_ary} 
\end{figure}

\subsubsection{\textbf{RE in the Scientific Field}}
In the scientific domain, in order to minimize the time invested in the scientific literature search, 
researchers have proposed methods to automatically extract the relations of scientific articles automatically.
\citeauthor{Augenstein2017SemEval2T} \shortcite{Augenstein2017SemEval2T} proposed the SemEval task for extracting keyphrases and the corresponding relations between them in the scientific texts.
\citeauthor{Luan2018InformationEF} \shortcite{Luan2018InformationEF} proposed a semi-supervised learning framework for scientific RE.
\citeauthor{Luan2018MultiTaskIO} \shortcite{Luan2018MultiTaskIO} developed a unified framework SciIE for extracting entities, relations, and coreferences in scientific documents.
\citeauthor{hou2019identification} \shortcite{hou2019identification} constructed a scientific leaderboard for extracting four items from NLP papers, including task, dataset, metric, and score. This benefits the community in keeping track of interesting scientific results.
\citeauthor{Eberts2019SpanbasedJE} \shortcite{Eberts2019SpanbasedJE} proposed a transformer-based joint RE model based on SciERC.
\citeauthor{jain2020scirex} \shortcite{jain2020scirex} 
created the SciREX dataset for the document-level N-ary relation extraction from scientific articles. 
As shown in Figure~\ref{fig:n_ary}, the key challenge is to detect the target triplets residing in multiple modalities, including paragraphs and tables of the document. 
\citeauthor{kruiper2020layman} \shortcite{kruiper2020layman} 
introduced the semi-open RE task to comprehend the most significant relationships governing the central concepts in the document. 
Different from previous works solely considering the content of the paper, CitaionIE further leverages the citation graph of referential links, showing the paper’s place in the broader literature. 
\citeauthor{Magnusson2021ExtractingFK} \shortcite{Magnusson2021ExtractingFK} built a SciClaim knowledge graph with entities, relations, and attributes. SciClaim contains both coarse-grained and fine-grained entity spans and relations from scientific claims.

\begin{table}[]
\caption{Two typical input and output examples of generation methods for relation extraction.}
\label{tab:generation_examples}
\small
\centering
\begin{tabular}{ p{1.5cm} | p{6cm}| p{5cm}}
\toprule
\textbf{Work}                                          & \textbf{Input Example}                                                                   & \textbf{Output Example}                                                                                                                    \\
\midrule
TANL~\cite{paolinistructured}     & Tolkien’s epic novel The Lord of the Rings was published in 1954-1955. & {[} Tolkien | person {]}'s epic novel {[} The Lord of the Rings | book | author = Tolkien {]} was published in 1954-1955 \\    \midrule
ChatIE~\cite{wei2023zero}     &   Given sentence: ``Japan then laid siege to the Syrian penalty area and had a goal disallowed for offside in the 16th minute.'' The known entity types are: {[}‘LOC’, ‘MISC’, ‘ORG’, ‘PER’{]}. Please answer: What types of entities are included in this sentence?  &    LOC, MISC \\
\bottomrule
\end{tabular}
\end{table}

\subsubsection{Prospects on Domain-specific RE}
Overall, we notice that: (1) extensive studies have focused on biomedical relation extraction. However, there is still a great demand for publicly available data resources and effective approaches in other specific domains. (2) Some domain-specific pre-trained language models have been proposed to address the lack of high-quality, large-scale labeled domain data. 
The corresponding PLMs injecting domain-specific information include BioBERT~\cite{lee2020biobert}, SciBERT \cite{beltagy2019scibert}, FinBERT \cite{YangKZZhangPKKannan2020FinBERTAP}, and Legal-BERT \cite{chalkidis2020legal}. These tasks are challenging due to the specialized vocabulary and the complexity of the relationships involved. Therefore, the continued advancement of relation extraction techniques specifically tailored to these domains is essential for various domain-specific applications. For example, 
\citeauthor{Roy2021IncorporatingMK} \shortcite{Roy2021IncorporatingMK} incorporated entity-level KG into pre-trained BERT for clinical RE, which integrates medical knowledge by several techniques.

\textbf{Summary}. 
Despite the remarkable progress made by previous works, there is still substantial room for improving the RE performance in specific domains.
(1) It is essential to further develop benchmark datasets and methods to identify and extract more practical and specific relations in different application domains. Current domain-specific datasets are either too narrow, containing only a small number of semantic relations, or too broad, containing an unbounded number of generic relations extracted from large and generic corpora \cite{kruiper2020layman}.
(2) PLMs have made a significant contribution to RE, and it would be a promising direction to further tailor domain-specific PLMs by injecting domain knowledge into the general PLMs to understand specialized vocabulary and tackle the complexity of the involved relations.

\section{RE with Pre-trained Language Models}
\label{sec:PLMs}

\begin{table*}[]
\caption{Examples of prompts for relation extraction in \cite{jiang2020can}.}
\label{tab:prompt_examples}
\centering
\begin{tabular}{clll}
\hline
\textbf{ID} & \multicolumn{1}{c}{\textbf{Relations}} & \multicolumn{1}{c}{\textbf{Manual Prompts}} & \multicolumn{1}{c}{\textbf{Mined Prompts}} \\ \hline
P140        & religion                               & x is affiliated with the y religion         & x who converted to y                       \\
P159        & headquarters location                  & The headquarter of x is in y                & x is based in y                            \\
P20         & place of death                         & x died in y                                 & x died at his home in y                    \\
P264        & record label                           & x is represented by music label y           & x recorded for y                           \\
P279        & subclass of                            & x is a subclass of y                        & x is a type of y                           \\
P39         & position held                          & x has the position of y                     & x is elected y                             \\ \hline
\end{tabular}
\end{table*}

Recently, PLMs have proved to be powerful in improving the performance of relation extraction \cite{Xu2022TowardsRL}, as demonstrated in Section~\ref{sec:encoder_PLMs}, where PLMs are used for context encoding.
As illustrated in Table~\ref{tab:performance_comparison}, the first part \cite{chen2022learning} of Table~\ref{tab:performance_comparison} shows the results of state-of-the-art RE approaches with fine-tuned PLMs on two benchmark datasets in the general domain (i.e., NYT and WebNLG). 
The second part \cite{gutierrez2022thinking} of Table~\ref{tab:performance_comparison} compares the few-shot performance between fine-tuned PLMs on two biomedical RE datasets, i.e., ChemProt and DDI. 
Specifically, three PLMs (i.e., PubMedBERT-base \cite{gu2021domain}, BioBERT-large \cite{lee2020biobert}, and RoBERTa-large \cite{liu2019roberta}) are fine-tuned on 100 labeled training samples. 
From Table~\ref{tab:performance_comparison}, we can observe that PLMs with more parameters usually outperform those with fewer parameters. Some well-designed BERT-based models produce competitive results compared to models with larger PLMs (i.e., BART and RoBERTa). 

Although PLMs have contributed significantly, supervised fine-tuning still suffers from a lack of sufficient supervised RE data in practice. 
In addition, there is a significant gap between the training objectives of the pre-training and fine-tuning processes in PLMs, which may hinder the adaptation of the knowledge in PLMs, especially for few-shot relation extraction (FSRE).
To overcome this limitation, prompt-tuning techniques \cite{Son2022GRASPGM} have been proposed to bridge the gap between pre-training and fine-tuning processes by converting downstream RE tasks into a language model format. 
This approach aims to leverage the capabilities of the PLM to perform a specific task by adapting it to the target task through training on a smaller, task-specific dataset.
The key idea is to reformulate the tasks by appending an instruction phase that can be directly solved by PLMs. 
Therefore, prompt learning casts the RE task as the text generation problem. This approach appends the templates to input sentences, introducing additional information into templates to aid the generation process. The prompts/templates appropriately define the relationship and order for the entity spans and labels. For example, the first example in table~\ref{tab:generation_examples},
\citeauthor{paolinistructured} \shortcite{paolinistructured} enclosed each entity and possibly some relations with special tokens [ ].  
The sequence of |-separated tags represents the entity type and a list of relations in the format “X = Y”, where X is the relation type, and Y is the tail entity of the relation. 
Besides, some recent developments \cite{jiang2020can, schick2021exploiting,han2021ptr} in the field of RE include the use of prompt-based approaches to prompt a PLM by converting the extraction of relation to predict the missing words. As shown in Table~\ref{tab:prompt_examples}, the mined prompts \cite{jiang2020can} are constructed from Wikipedia through both middle words and dependency paths. The manual prompts are created by experts according to the relation semantics, which is more complicated syntactically.
However, manually defining the appropriate mapping phrase is time-consuming and non-intuitive \cite{jiang2020can} since it requires task-specific knowledge and manual identification words that the PLM can sufficiently understand.

\begin{table*}
\small
 \caption{An overview performance comparison of RE methods with PLMs in general and specific domains.}
    \label{tab:performance_comparison}
	\centering
		% {\renewcommand{\arraystretch}{1.0}
			% \resizebox{1.0\columnwidth}{!}{
            % \begin{tabular}{ p{1.2cm} | p{1.4cm}| p{2cm} | p{4.5cm} | p{1.2cm}| p{3.4cm} }
            \begin{tabular}{lcllllll}
                \hline
                                                                   & \multicolumn{1}{l|}{}              & \multicolumn{3}{c|}{\textbf{NYT}}               & \multicolumn{3}{c}{\textbf{WebNLG}} \\ \hline
                \multicolumn{1}{l|}{\textbf{Method}}                        & \multicolumn{1}{l|}{\textbf{\# PLM Param.}} & \textbf{Prec.} & \textbf{Rec.} & \multicolumn{1}{l|}{\textbf{F1}} & \textbf{Prec.}    & \textbf{Rec.}   & \textbf{F1}     \\ \hline
                \multicolumn{8}{c}{\textbf{RE methods with PLMs (In general domain)}}                                                                                                                                 \\ \hline
                CasRel~\cite{Wei2019ANC}          & BERT$_{(110\text{M})}$                         & 89.7  & 89.5 & 89.6                    & 93.4     & 90.1   & 91.7   \\
                TPLinker~\cite{Wang2020TPLinkerSJ} & BERT$_{(110\text{M})}$                          & 91.3  & 92.5 & 91.9                    & 91.8     & 92.0   & 91.9   \\
                CGT~\cite{ye2021contrastive}       & UniLM$_{(110\text{M})}$                         & 94.7  & 84.2 & 89.1                    & 92.9     & 75.6   & 83.4   \\
                PRGC~\cite{zheng2021prgc}          & BERT$_{(110\text{M})}$                          & 93.3  & 91.9 & 92.6                    & 94.0     & 92.1   & 93.0   \\
                REBEL~\cite{cabot2021rebel}        & BART$_{(406\text{M})}$                        & -     & -    & 93.4                    & -        & -      & -      \\
                R-BPtrNet~\cite{chen2021jointly}   & RoBERTa$_{(335\text{M})}$                      & 94.0  & 92.9 & 93.5                    & 94.3     & 93.3   & 93.8   \\
                MTG~\cite{Chen2022LearningRP}                                                & T5-large$_{(770\text{M})}$                     & 95.6  & 93.1 & 94.3                    & 94.8     & 95.1   & 94.9   \\ \hline
                                                                   & \multicolumn{1}{l|}{}              & \multicolumn{3}{c|}{\textbf{ChemProt}}          & \multicolumn{3}{c}{\textbf{DDI}}    \\ \hline
                \multicolumn{1}{l|}{\textbf{Method}}                        & \multicolumn{1}{l|}{\textbf{\# PLM Param.}} & \textbf{Prec.} & \textbf{Rec.} & \multicolumn{1}{l|}{\textbf{F1}} & \textbf{Prec.}    & \textbf{Rec.}   & \textbf{F1}     \\ \hline
                \multicolumn{8}{c}{\textbf{RE methods with PLMs (In specific domain)}}                                                                                                                                 \\ \hline
                PubMedBERT-Base                                    & 100M                               & 17.9  & 62.0 & 27.7                    & 19.9     & 79.1   & 31.8   \\
                BioBERT-Large                                      & 345M                               & 19.0  & 60.6 & 28.7                    & 17.3     & 75.4   & 28.2   \\
                RoBERTa-Large                                      & 354M                               & 22.0  & 69.7 & 33.4                    & 25.5     & 77.9   & 38.4   \\
                % GPT-3   In-Context                                 & 175B                               & 15.9  & 68.9 & 25.9                    & 9.6      & 48.6   & 16.1   \\ 
                \hline
            \end{tabular}
	% }}
\end{table*}

To avoid the labor-intensive process of constructing prompts, recent works \cite{shin2020autoprompt,gao2021making,schick2020automatically} pay attention to automatically generating and searching prompts. For example, \citeauthor{shin2020autoprompt} \shortcite{shin2020autoprompt} designed AUTOPROMPT to automatically create prompts by a gradient-guided search. It shows that masked language models (MLMs) can be effectively used as relation extractors without additional fine-tuning.
Moreover, some studies \cite{li2021prefix,qin2021learning,lester2021power} propose continuous prompts while fixing all PLM parameters, and experiments show that such soft prompts work well on few-shot RE datasets. 
Drawing inspiration from prompting, \citeauthor{li2021prefix} \shortcite{li2021prefix} proposed the prefix-tuning, which attends the subsequence tokens to prompt the PLMs. 
\citeauthor{qin2021learning} \shortcite{qin2021learning} 
 and \citeauthor{lester2021power} \shortcite{lester2021power} proposed to model prompts as continuous vectors optimized by a mixture of prompts.

What's more, recent advances \cite{xu2023unleash,sun2024consistency,jiang2024genres} in large language models (LLMs), such as GPT-3 \cite{brown2020language}, ChatGPT and GPT-4\footnote{The corresponding model versions of GPT-3, ChatGPT and GPT-4 are text-DaVinci-003, GPT-3.5-turbo, and GPT-4-turbo, respectively.} \cite{achiam2023gpt}, have demonstrated their exceptional performance across various natural language processing (NLP) tasks. 
While PLMs primarily strive for high performance in predefined NLP tasks, LLMs exhibit emergent capabilities extending beyond task-specific learning. GPT-3 represents a significant milestone in the evolution from PLMs to LLMs~\cite{zhao2023survey}. 
With the continuous growth in model parameters and training corpus size, LLMs exhibit emergent abilities that enable them to engage in in-context learning (ICL), where the models can reason from a small number of demonstration examples within the input context \cite{dong2022survey}. 
For example, in the second example in Table~\ref{tab:generation_examples}, LLMs can effectively perform relation extraction given specific prompts \cite{wei2023zero}. 
Besides, some LLM-based methods \cite{jiang2024genres,agrawal2022large,wei2023zero} also provide several example demonstrations in the input, fully taking advantage of LLMs' larger number of parameters and longer input context lengths. 
\citeauthor{jiang2024genres} \shortcite{jiang2024genres} tested the capabilities of the leading LLMs to perform RE in a zero-shot manner, which includes the GPT Family \cite{brown2020language}, i.e., text-davinci-003, gpt-3.5-turbo, gpt-3.5-turbo-instruct and gpt-4-turbo \cite{achiam2023gpt}, and the LLaMA family \cite{touvron2023llama}, i.e., LLaMA-2-7B, LLaMA-2-70B, Vicuna-1.5-7B, Vicuna-1.3-33B, and WizardLM-70B \cite{xu2023wizardlm}. 
\citeauthor{agrawal2022large} \shortcite{agrawal2022large} showed that LLMs perform well at zero- and few-shot clinical relation extraction despite not being trained specifically in clinical texts. The construction of demonstrations facilitates the LLMs' comprehension and easy answer extraction.
\citeauthor{wei2023zero} (\shortcite{wei2023zero}) explored the helpfulness of ChatGPT in the RE task and proposed a two-stage framework (ChatIE). This framework transforms the zero-shot IE task into a multi-turn question-answering problem by prompting ChatGPT and improves the experimental results.

\textbf{Summary}. 
RE tasks have benefited significantly from both PLMs and LLMs. PLMs have shown remarkable performance in improving relation extraction by leveraging pre-training on large corpora, especially for fine-tuning PLMs for specific RE tasks. However, existing methods relying solely on PLMs often face challenges when dealing with newly emerging relations due to the need for extensive data annotation, which can be time-consuming and labor-intensive. 
LLMs showcase impressive capabilities in generation and have inspired exploration of alternative approaches for obtaining auto-labeled documents with new relations. They excel in scenarios with limited annotations, where their memorization and reasoning capabilities contribute significantly to relation extraction tasks. However, the inference latency and financial cost associated with calling LLMs' APIs are higher compared to fine-tuning PLMs.

\begin{figure*}[t] 
	\centering 
	\includegraphics[width=0.9 \textwidth]{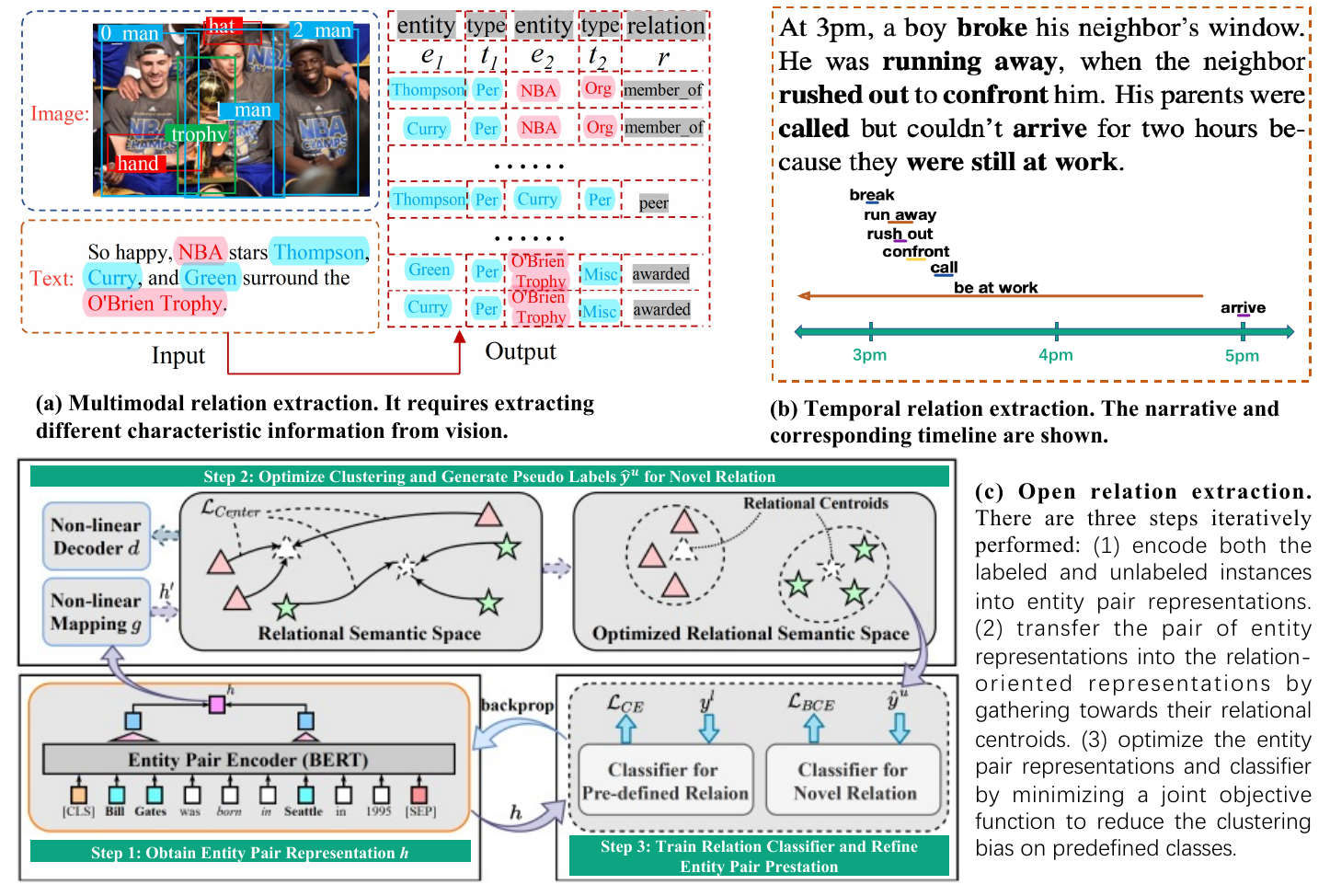} 
	\caption{Illustration of the future directions mentioned in Section~\ref{sec:future_directions}. }
	\label{fig:future_directions} 
\end{figure*}

\section{Future Directions}
\label{sec:future_directions} 
RE studies have made significant progress in recent years regarding new neural RE model designs and subtasks. 
However, challenges and limitations remain that need to be addressed, including the need for more diverse data in practical scenarios, handling complex and unevenly distributed relations, and incorporating additional new relation types.

\subsection{Multi-modal Relation Extraction}
Along with text, images, and videos have become popular ways to convey information on the internet. 
This highlights the importance of extracting relations from multi-modal input rather than textual data alone. 
Multi-modal RE takes advantage of the large visual-text corpus by focusing on extracting relations from these media forms.
\citeauthor{zheng2021mnre} \shortcite{zheng2021mnre}
proposed the multi-modal RE dataset MNRE containing visual evidence collected from social media posts. Subsequently, 
\citeauthor{zheng2021multimodal} \shortcite{zheng2021multimodal} 
introduced the multi-modal RE model to capture the knowledge from related information in the texts and images.
However, many interesting problems remain to be explored \cite{wang2022named}. As shown in Figure~\ref{fig:future_directions}(a) \cite{yuan2022joint}, the multi-modal RE task takes images and texts as input, then recognizes the entities and corresponding relations from the multi-modal data. This task is expected to align the entity-entity relations in the text with the object-object relations in the images. 
Since multi-modal data is often closely related, the visual content can supplement missing semantics of the textual content and improve the performance of RE methods. Thus, it is crucial to develop well-constructed multi-modal RE methods combining visual and textual information to extract relations more accurately.

\begin{table*}[]
\small
\caption{The number of sentences and relations for each language shown in the multi-lingual RE dataset \cite{seganti2021multilingual}.}
\label{tab:future_1_directions} 
\centering
\begin{tabular}{cccccccccccccccc}
\toprule
                   & EN   & KO  & IT  & FR  & DE  & PT  & NL  & PL  & ES  & AR & RU & SV & FA & UK & \textbf{Total} \\ \midrule
\textbf{Sentences} & 748k & 20k & 76k & 62k & 53k & 45k & 40k & 17k & 12k & 9k & 7k & 5k & 3k & 1k & \textbf{1.1M}  \\ \midrule
\textbf{Relations} & 36   & 28  & 22  & 22  & 22  & 22  & 22  & 22  & 22  & 9  & 8  & 22 & 8  & 7  & \textbf{36}    \\ \bottomrule
\end{tabular}
\end{table*}

\subsection{Cross-lingual Relation Extraction}
Existing state-of-the-art RE systems are primarily available for English because they heavily rely on annotated corpora and PLMs. 
These methods perform RE on a sentence in a source language by first translating it into English, then performing RE on the translated sentence, and finally projecting the identified phrase back to the source language.
However, these methods assume that parallel bilingual corpora can be obtained by existing machine translation systems.
It is challenging to mitigate the noisy data problem caused by machine translation systems and align the sentences and extracted triples between different languages. 
One future research direction is to explore the \textit{cross-lingual projection methods} for language-independent RE \cite{faruqui2015multilingual}. 
Therefore, some works have been proposed to improve cross-lingual transfer for RE, including utilizing universal dependency structure parses \cite{subburathinam2019cross} and mBERT \cite{rathore2022pare}.
Recent progress \cite{Bhartiya2021DiSReXAM,seganti2021multilingual} demonstrates that multi-lingual training can improve performance across all languages in RE since the relation information from other languages might help encode the information in a given language.
These methods learn language-agnostic sentence representations in complex and multi-lingual common spaces. 
As shown in Table~\ref{tab:future_1_directions}~\cite{seganti2021multilingual}, the distribution for each language is quite different. One of the main challenges in cross-lingual relation extraction is dealing with language differences. Languages vary in their grammatical structure, vocabulary, and syntax, which makes it difficult to identify relationships between entities across languages. Another challenge is the ambiguity of words and their translations across languages. To tackle these challenges, it is crucial to investigate diverse approaches for aligning relation semantics between resource-rich languages and those with more limited data availability.

\subsection{Temporal Relation Extraction}
\label{temporal_RE}
Temporal RE aims to identify relations between entities subject to temporal constraints, enhancing the applicability of RE systems in complex reasoning.
As shown in Figure~\ref{fig
}(b) \cite{vashishtha2019fine}, a timeline illustrates the fine-grained (real-valued) temporal relations implicated in the text, mapping the temporal relations and event durations to real-valued scales.
There are two mainstream approaches dedicated to temporal RE: dealing with relations between events and time expressions \cite{tan2021extracting} and extracting relations between entities at a given time spot through temporal reasoning \cite{yan2019relation}.
Although previous studies have attempted to address this issue by generating patterns for time-variant relations, many challenges remain, including (i) the complex dependencies between entities, relations, and conditions; (ii) the difficulty of handling conditions in various forms in free text; and (iii) the lack of well-annotated data.
Therefore, a general framework is needed to formalize the conditional dimensions.

\subsection{Evolutionary Relation Extraction}
\label{evolutionary_RE}
In recent years, most RE paradigms have been designed on pre-defined relation sets. 
However, as our world experiences continuous expansion of new relations, it is infeasible for RE systems to handle all emerging relation types. 
Therefore, there is a demand for RE systems that can generalize to new relations beyond pre-defined schemes.
Several works have been proposed to handle new relations, which mainly fall into two groups: (1) \textbf{open relation extraction}. As illustrated in Figure~\ref{fig:future_directions}(c) \cite{zhao2021relation}, open information extraction approaches \cite{hu2020selfore} extract related phrases as representations of relations and entities from the text. Another type is the clustering-based unsupervised relation discovery method \cite{zhao2021relation}, which discovers unseen relation types using clustering optimization;
(2) \textbf{lifelong relation extraction}. This group of methods \cite{zhao2022consistent} aims to continuously train a RE model to learn new relations while avoiding forgetting the accurate classification of old ones. 
Evolutionary RE is a promising research area, giving RE models the ability to generalize beyond the training data and learn from new data.
However, many unresolved challenges remain. For open relation extraction, where phrases of the same relation can have various forms, the key challenge is to canonicalize relation phrases to reduce ambiguity and redundancy. For lifelong RE, more efforts are needed to prevent RE models from overfitting the experience memory.
It is worth exploring more effective methods leveraging large language models (as discussed in Section~\ref{sec:PLMs}) to tackle the challenges in evolutionary RE effectively.

\subsection{Explainable Relation Extraction}
Despite significant advancements in RE over the past decade, the opacity of DL-based RE models has led to an increasing demand for explainability. 
The core challenge in achieving explainability lies in the intrinsic complexity of RE models \cite{ribeiro2016should}, which often function as black boxes. Another obstacle is that the features extracted by RE models may not be directly interpretable by humans. 
This disconnect complicates efforts to comprehend the underlying rationale behind the model's decisions \cite{saeed2023explainable}, obscuring the reliability and accuracy of the extracted relations. Such opacity hampers users' ability to trust the models. 
To address these challenges, future research needs to focus on developing methods that provide accurate, real-time explanations of model predictions, particularly shedding light on how these models arrive at their conclusions. 
By enhancing explainability, the RE models could advance in their capabilities and become more trustworthy, enabling broader adoption in critical domains where transparency and reliability are essential.

\section{Conclusion}
This survey provided an up-to-date and comprehensive review of recent advances in relation extraction. 
We first designed a novel taxonomy to systematically summarize the model architectures used in existing DNN-based RE approaches, fully combing recent research trends in categories, and illustrating the differences and connections between RE subtasks. Then, we analyzed several important yet challenging RE problems and their corresponding solutions. Specifically, we discussed the performance of relation extraction on current solutions in diverse, challenging settings (i.e., the low-resource setting and the cross-sentence setting) and specific domains (i.e., biomedical, finance, legal, and scientific fields).
Considering the new frontiers in RE studies, we also presented in-depth analyses that revealed the issues of RE with PLMs and LLMs. Finally, we pointed out several promising future directions and prospects.
We hope this survey provides insightful perspectives and inspires the widespread implementation of real-life RE systems.

\section*{Acknowledgments}
This research is supported in part by grants from the Research Grant Council of the Hong Kong Special Administrative Region, China (No. CUHK 14217622).
Min Yang was supported by National Key Research and Development Program of China (2022YFF0902100), National Natural Science Foundation of China (62376262), the Natural Science Foundation of Guangdong Province of China (2024A1515030166), Shenzhen Science and Technology Innovation Program (KQTD20190929172835662), Shenzhen Basic Research Foundation (JCYJ20210324115614039).

\bibliographystyle{ACM-Reference-Format}
\bibliography{sample-acmsmall}

% \appendix

\end{document}